%% file: main.tex
\documentclass[10pt,twocolumn,letterpaper]{article}

\usepackage[pagenumbers]{cvpr}

\definecolor{cvprblue}{rgb}{0.21,0.49,0.74}
\usepackage[pagebackref,breaklinks,colorlinks,allcolors=cvprblue]{hyperref}

\input{preamble}

\def\paperID{*****}
\def\confName{CVPR}
\def\confYear{2026}

\title{OVGGT: $O(1)$ Constant-Cost Streaming Visual Geometry Transformer}

\author{
Si-Yu Lu\textsuperscript{1}\orcidlink{0009-0003-5282-8977} \quad
Po-Ting Chen\textsuperscript{2}\orcidlink{0009-0001-9958-0994} \quad
Hui-Che Hsu\textsuperscript{2}\orcidlink{0009-0006-1607-6319} \quad
Sin-Ye Jhong\textsuperscript{2}\orcidlink{0000-0003-4481-1633} \\
Wen-Huang Cheng\textsuperscript{1}\orcidlink{0000-0002-4662-7875} \quad
Yung-Yao Chen\textsuperscript{2}\orcidlink{0000-0001-6852-8862} \\[0.5em]
\textsuperscript{1}National Taiwan University \quad
\textsuperscript{2}National Taiwan University of Science and Technology
}

\begin{document}

% Patch \@maketitle to include teaser in the full-width title area
\makeatletter
\let\old@maketitle\@maketitle
\def\@maketitle{%
  \old@maketitle
  \vspace{-0.5em}
  \begin{center}
    \includegraphics[width=\textwidth]{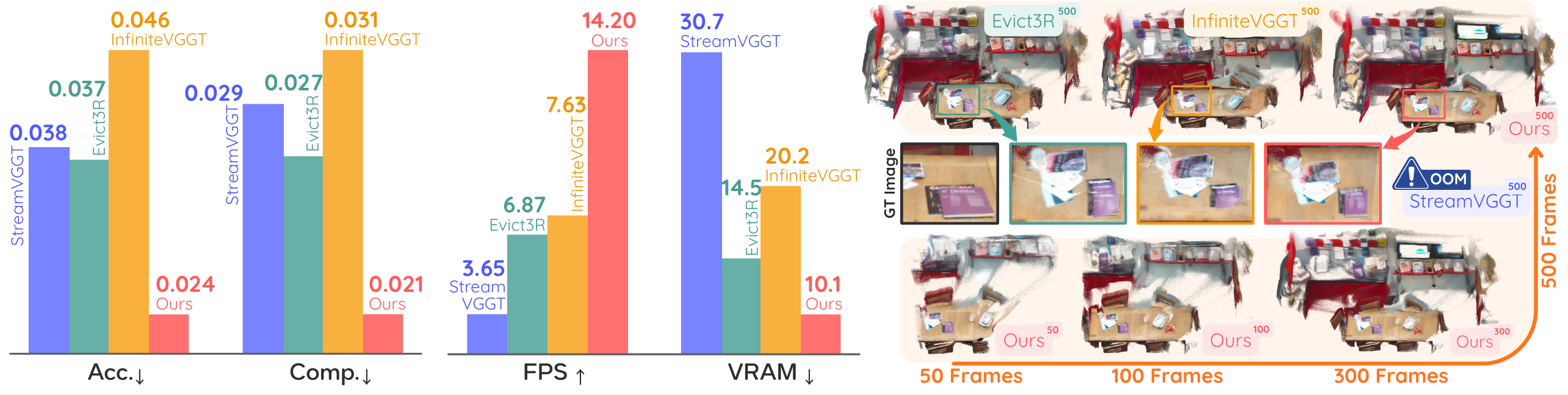}
    \captionof{figure}{\textbf{Streaming 3D on a single 32\,GB GPU.}
    \textbf{Left:} On 200-frame sequences~\cite{shotton2013scenes}, \model{} outperforms all baselines in reconstruction quality, speed, and VRAM usage.
    \textbf{Right:} From 50 to 500 frames, StreamVGGT runs out of memory; other methods survive but suffer notable quality degradation. \model{} maintains high-fidelity reconstructions at lower cost.}
    \label{fig:teaser}
  \end{center}
  \vspace{0.5em}
}
\makeatother

\maketitle

% ---------------------------------------------------------------
% Paper content

\input{sec/0_abstract}

\input{sec/1_intro}

\input{sec/2_related}
\input{sec/3_method}
\input{sec/4_exp}

\input{sec/8_conclusion}

% ---------------------------------------------------------------
% Bibliography
{
    \small
    \bibliographystyle{ieeenat_fullname}
    \bibliography{main}
}

\input{sec/99_supp}

\end{document}

%% file: preamble.tex
\usepackage{graphicx}
\usepackage{stfloats}
\usepackage{booktabs}
\usepackage{multirow}
\usepackage{amssymb}
\usepackage{orcidlink}

\graphicspath{{figure/}}

\newcommand{\model}{OVGGT}

\usepackage{xspace}
\makeatletter
\DeclareRobustCommand\onedot{\futurelet\@let@token\@onedot}
\def\@onedot{\ifx\@let@token.\else.\null\fi\xspace}
\def\eg{\emph{e.g}\onedot}

\makeatother

%% file: sec/0_abstract.tex
% !TEX root = ../main.tex

\begin{abstract}

Reconstructing 3D geometry from streaming video requires continuous inference under bounded resources.
Recent geometric foundation models achieve impressive reconstruction quality through all-to-all attention, yet their quadratic cost confines them to short, offline sequences.
Causal-attention variants such as StreamVGGT enable single-pass streaming but accumulate an ever-growing KV cache, exhausting GPU memory within hundreds of frames and precluding the long-horizon deployment that motivates streaming inference in the first place.
We present \model{}, a training-free framework that bounds both memory and compute to a fixed budget regardless of sequence length.
Our approach combines Self-Selective Caching, which leverages FFN residual magnitudes to compress the KV cache while remaining fully compatible with FlashAttention, with Dynamic Anchor Protection, which shields coordinate-critical tokens from eviction to suppress geometric drift over extended trajectories.
Extensive experiments on indoor, outdoor, and ultra-long sequence benchmarks demonstrate that \model{} processes arbitrarily long videos within a constant VRAM envelope while achieving state-of-the-art 3D geometric accuracy.

\end{abstract}

%% file: sec/1_intro.tex
% !TEX root = ../main.tex

\section{Introduction}
\label{sec:intro}

Reconstructing 3D scene geometry from sequential image observations is a cornerstone problem in computer vision, underpinning autonomous navigation, augmented reality, robotic manipulation, and large-scale digital twin construction.
The task requires inferring dense, metrically consistent spatial structures from 2D image streams, reconciling the inherent ambiguity of monocular observations with multi-view geometric constraints.
For decades, this progress has been driven by classical pipelines that decompose this problem into cascaded stages: keypoint detection and matching~\cite{detone2018superpoint,sarlin2020superglue,lindenberger2023lightglue}, robust pose estimation, triangulation, and bundle adjustment~\cite{schoenberger2016sfm}.
While effective under controlled conditions, these modular designs are inherently fragile: errors in any single stage propagate downstream, limiting robustness on textureless surfaces, repetitive structures, or large viewpoint changes.

DUSt3R~\cite{wang2024dust3r} marked a paradigm shift, ushering in the era of Geometric Foundation Models by training a single Transformer network to regress dense 3D pointmaps from image pairs end-to-end, bypassing the multi-stage pipeline entirely without requiring camera intrinsics or explicit feature matching.
Follow-up works augmented this framework with dense correspondences~\cite{leroy2024mast3r} and dynamic-scene support~\cite{zhang2024monst3r}.
Yet the pairwise nature fundamentally limits scalability: extending to $N$ views demands $O(N^2)$ predictions followed by costly global alignment optimization.
While subsequent streaming variants~\cite{wang2024spann3r,wang2025cut3r} achieve continuous inference through dedicated architectural designs, they still suffer from accuracy degradation over long input sequences.

To circumvent costly global alignment, subsequent works~\cite{wang2025vggt,yang2025fast3r,yang2024mvdust3rplus} sought to advance the paradigm through all-to-all attention designs.
For instance, VGGT~\cite{wang2025vggt} jointly processes all views through alternating intra-frame and global all-to-all attention, predicting cameras, depth, and point clouds in one forward pass.
Nevertheless, the quadratic cost of attention persists: VGGT exhausts 80\,GB of GPU memory at merely ${\sim}300$ frames, and the paradigm inherently precludes continuous inference, as every invocation must recompute over all previous inputs.
To address this limitation, StreamVGGT~\cite{zheng2025streamvggt} reformulated the architecture into temporal causal attention akin to autoregressive decoding, caching all prior KV pairs so that each frame is processed exactly once, thereby enabling streaming inference without redundant recomputation at each time step.
However, the linear growth of the KV cache remains a critical bottleneck: 100 frames already accumulate over $10^5$ tokens per layer (${\sim}10$\,GB VRAM), and the per-step attention cost escalates with sequence length, fundamentally preventing deployment on the long sequences demanded by streaming 3D reconstruction.

In this work, we present \model{}, a geometric foundation model for online streaming that maintains constant memory and compute regardless of sequence length, built upon two complementary components: {Self-Selective Caching (SSC)} and {Dynamic Anchor Protection (DAP)}.
SSC compresses the inference-time cache to a fixed budget via (i)~\emph{Activation Value Rating}, which scores each token's geometric salience using its FFN residual magnitude, a quantity already computed in the forward pass and fully compatible with FlashAttention~\cite{dao2022flashattention,dao2024flashattention2}, with spatial Gaussian smoothing to encourage coherent retention; and (ii)~\emph{Cache Compression}, which unifies current-frame activation scores with historical key-vector diversity to balance geometric importance and distributional coverage.
To maintain geometric stability over long sequences, DAP shields two types of anchors from eviction: a \emph{Global Initial Anchor} that permanently protects all first-frame tokens to preserve coordinate-system consistency, and \emph{Historical Anchors} that are adaptively registered based on view-overlap coverage to supply long-range geometric references.
Both components are entirely training-free, requiring no architectural modifications and readily applicable as a plug-in for pretrained causal-attention models.
Experiments show that \model{} processes arbitrarily long sequences within a fixed VRAM envelope while surpassing the full-cache StreamVGGT in reconstruction accuracy.

In summary, our contributions are as follows.
\textbf{(1)}~We present \model{}, a training-free online streaming framework that performs 3D inference from arbitrarily long videos under fixed memory and compute, eliminating the scaling bottleneck of existing causal-attention pipelines.
\textbf{(2)}~We design Self-Selective Caching, combining FFN-residual-based Activation Value Rating with spatial smoothing and Hybrid Scoring to compress the KV cache to a fixed budget while remaining fully compatible with FlashAttention.
\textbf{(3)}~We introduce Dynamic Anchor Protection, which shields coordinate-critical tokens from eviction through a global initial anchor and historical anchors, effectively suppressing geometric drift over extended trajectories.
Extensive experiments demonstrate state-of-the-art geometric accuracy on indoor, outdoor, and ultra-long sequence benchmarks, with faster throughput and lower memory consumption than existing causal streaming methods.

%% file: sec/2_related.tex
% !TEX root = ../main.tex
% \newpage

\section{Related Work}
\label{sec:related}

\subsection{Classical Geometric Reconstruction}
\label{sec:related:classical}

\noindent\textbf{Structure from Motion.}
Structure-from-Motion (SfM) recovers camera poses and sparse 3D point clouds from unordered images via feature extraction, pairwise matching, and joint optimization.
COLMAP~\cite{schoenberger2016sfm,schoenberger2016mvs} remains the dominant incremental system, and recent learned matchers~\cite{detone2018superpoint,sarlin2020superglue,lindenberger2023lightglue} and global SfM methods~\cite{pan2024glomap} have improved robustness and runtime, while fully differentiable pipelines~\cite{wang2024vggsfm} enable end-to-end learned reconstruction.
Nevertheless, the reliance on explicit correspondences and iterative optimization fundamentally limits throughput, making SfM difficult to deploy in real-time streaming scenarios and inherently unable to produce dense reconstructions.

\noindent\textbf{Multi-View Stereo.}
Multi-View Stereo (MVS) methods densify SfM's sparse output into complete surface models.
Classical approaches such as PMVS~\cite{furukawa2010pmvs} employ patch-based matching, while learning-based methods~\cite{yao2018mvsnet,gu2020cascade,wang2021patchmatchnet} have progressively replaced hand-crafted components with end-to-end differentiable cost volumes.
Despite considerable progress, MVS methods typically assume known poses and intrinsics and operate offline, precluding streaming reconstruction from uncalibrated video.

\noindent\textbf{Simultaneous Localization and Mapping.}
SLAM systems jointly estimate camera trajectory and scene structure in real time.
Feature-based methods such as ORB-SLAM~\cite{mur2015orbslam,mur2017orbslam2,campos2021orbslam3} track sparse landmarks with loop closure for global consistency, while learning-based approaches~\cite{teed2021droidslam,teed2024dpvo,lipson2024dpvslam} achieve superior accuracy through differentiable optimization.
However, both classical and learned SLAM systems typically yield only sparse or semi-dense maps, and their overall pipelines still operate in disjoint stages, precluding end-to-end optimization and limiting generalization.

\subsection{Geometric Foundation Models}
\label{sec:related:foundation}
Driven by large-scale pretraining and Vision Transformer architectures, geometric foundation models have emerged as a unified paradigm that jointly addresses pose estimation, depth prediction, and dense 3D reconstruction within a single feed-forward framework.
These models can be broadly categorized by how they aggregate multi-view information.

\noindent\textbf{Pairwise Prediction with Post-hoc Alignment.}
DUSt3R~\cite{wang2024dust3r} established the foundational paradigm by training a siamese Vision Transformer to directly regress per-pixel 3D pointmaps from an image pair, eliminating the need for camera intrinsics.
MASt3R~\cite{leroy2024mast3r} augmented the architecture with a dense feature head for 3D-grounded correspondence matching, and subsequent extensions addressed dynamic scenes~\cite{zhang2024monst3r}, real-time dense SLAM~\cite{murai2024mast3rslam}, and view symmetrization~\cite{cabon2025must3r}.
However, scaling to many views still necessitates $O(N^2)$ pairwise computations followed by global alignment~\cite{wang2024dust3r,leroy2024mast3r}, severely impeding deployment on long sequences.

\noindent\textbf{Offline All-to-All Attention.}
To circumvent pairwise scalability bottlenecks, VGGT~\cite{wang2025vggt} alternates intra-frame spatial self-attention with global cross-frame attention to predict cameras, depth, and point clouds in a single forward pass.
Fast3R~\cite{yang2025fast3r} scaled a similar architecture to over 1{,}000 images via FlashAttention and tensor parallelism.
Despite remarkable quality, the quadratic memory complexity of global attention confines these methods to offline batch processing.

\begin{figure*}[t!]
  \centering
  \includegraphics[width=\textwidth]{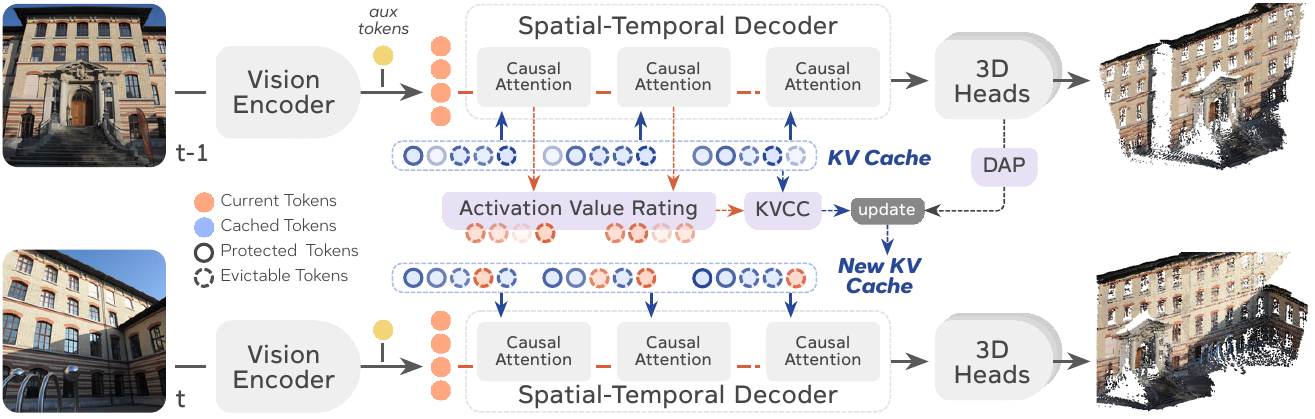}
  \caption{\textbf{Overview of \model{}.} At each time step, the input frame is encoded into tokens and processed by a spatial-temporal decoder that attends to a bounded KV cache. During inference, the Activation Value Rating module scores each token's geometric salience, and the KV Cache Compression (KVCC) module evicts low-scoring tokens to maintain a fixed cache budget. Dynamic Anchor Protection (DAP) shields coordinate-critical tokens from eviction, ensuring long-range geometric stability.}
  \label{fig:pipeline}
\end{figure*}

\noindent\textbf{Streaming Methods for Continuous Inference.}
To bridge representational capacity and unbounded video demands, Spann3R~\cite{wang2024spann3r} extended pairwise prediction to sequential processing with an external spatial memory.
CUT3R~\cite{wang2025cut3r} achieves constant-resource inference via a fixed-size recurrent state, and its follow-up TTT3R~\cite{chen2025ttt3r} further stabilizes long-sequence accuracy.
Point3R~\cite{wang2025point3r} leverages spatial pointer memory with hierarchical positional embeddings for cross-frame geometric aggregation.
However, these methods remain constrained in model capacity for capturing long-range dependencies.
Capitalizing on VGGT's superior accuracy, StreamVGGT~\cite{zheng2025streamvggt} converted its bidirectional attention into temporal causal attention with a KV cache, preserving much of the all-to-all representational capacity while enabling single-pass streaming.
However, the linearly growing KV cache remains the critical bottleneck: memory and latency increase monotonically with processed frames, eventually exceeding GPU capacity.
Current works~\cite{mahdi2025evict3r,yuan2026infinitevggt} also attempt to mitigate unbounded cache growth but still face imprecise resource control or degraded long-sequence accuracy.
Our work directly addresses this bottleneck by introducing fixed-budget cache management and dynamic anchoring, enabling long sequences under constant resource consumption while preserving geometric accuracy.

%% file: sec/3_method.tex
% !TEX root = ../main.tex
% \newpage

\section{Method}
\label{sec:method}

As illustrated in \cref{fig:pipeline}, \model{} is a streaming model that performs geometric inference from arbitrarily long video sequences under a fixed budget.
Based on the causal attention framework of StreamVGGT~\cite{zheng2025streamvggt}, it introduces token-level cache management and an anchoring mechanism that together enable continuous processing of thousands of frames within a constant VRAM envelope. 
We first analyze the bottleneck of the existing streaming model (\cref{sec:bottleneck}), then present Self-Selective Caching (\cref{sec:ssc}) and Dynamic Anchor Protection (\cref{sec:dap}).

% -----------------------------------------------------------------------------
\subsection{Preliminaries and Bottlenecks}
\label{sec:bottleneck}
% -----------------------------------------------------------------------------

StreamVGGT~\cite{zheng2025streamvggt} converts the offline all-to-all attention of VGGT~\cite{wang2025vggt} into a causal streaming pipeline for real-time 3D inference.
Each input frame $I_t$ is encoded by a frozen DINOv2~\cite{oquab2024dinov2} backbone into $N_p = \lfloor H/p \rfloor \times \lfloor W/p \rfloor$ patch tokens, which are concatenated with one learnable camera token $\mathbf{z}_{\mathrm{cam}} \in \mathbb{R}^{C}$ and four register tokens $\mathbf{z}_{\mathrm{reg}} \in \mathbb{R}^{4 \times C}$ (collectively termed \emph{aux tokens}), yielding a per-frame sequence of $M = 1 + 4 + N_p$ tokens.
These tokens pass through $L{=}24$ alternating attention blocks, each consisting of intra-frame spatial self-attention $\mathrm{SA}^{(l)}$ followed by cross-frame temporal causal attention $\mathrm{CA}^{(l)}$ that queries the KV cache $\mathcal{C}_{t}$.
The output tokens are decoded by camera, depth, and point cloud heads into per-frame camera parameters $\mathbf{g}_t \in \mathbb{R}^9$, depth map $\mathbf{D}_t \in \mathbb{R}^{H \times W}$, and 3D pointmap $\mathbf{P}_t \in \mathbb{R}^{H \times W \times 3}$:
\begin{equation}
  (\mathbf{g}_t,\, \mathbf{D}_t,\, \mathbf{P}_t)
  = \Phi\!\Bigl(
      \bigl[\mathrm{CA}^{(l)}\!\bigl(
        \mathrm{SA}^{(l)}(\cdot),\;
        \mathcal{C}_{t}^{(l)}
      \bigr)\bigr]_{l=0}^{L-1}
    \Bigr),
\end{equation}
where $\Phi$ subsumes the three prediction heads.
The KV cache grows by $M$ entries per layer per frame:
\begin{equation}
  \mathcal{C}_t^{(l)}
  = \bigl[\,\mathcal{C}_{t-1}^{(l)};\;
      (\mathbf{K}_t^{(l)},\, \mathbf{V}_t^{(l)})\,\bigr],
  \quad l = 0, \ldots, L{-}1.
\end{equation}
After $T$ frames, the total footprint is $\mathrm{Mem}(\mathcal{C}_T) = 2 \cdot L \cdot T \cdot M \cdot N_h \cdot d$, where $N_h$ and $d$ are the head count and the per-head dimension, respectively.
At $518{\times}392$ resolution ($M{=}1{,}041$), merely 100 frames consume ${\sim}10$\,GB of VRAM.
This linear growth imposes two bottlenecks: (i)~VRAM exhaustion caps the maximum sequence length, and (ii)~the per-step attention cost $O(M \cdot |\mathcal{C}_t^{(l)}|)$ increases with $t$, degrading throughput over time.

We address these issues by bounding the cache to a fixed budget $B$, which reduces the per-step cost to $O(M \cdot B)$, constant with respect to sequence length, thereby achieving $O(1)$ per-step inference and storage overhead.
The following subsections detail how we achieve this while preserving reconstruction quality.

% -----------------------------------------------------------------------------
\subsection{Self-Selective Caching}
\label{sec:ssc}
% -----------------------------------------------------------------------------

\begin{figure}[t]
  \centering
  \includegraphics[width=\columnwidth]{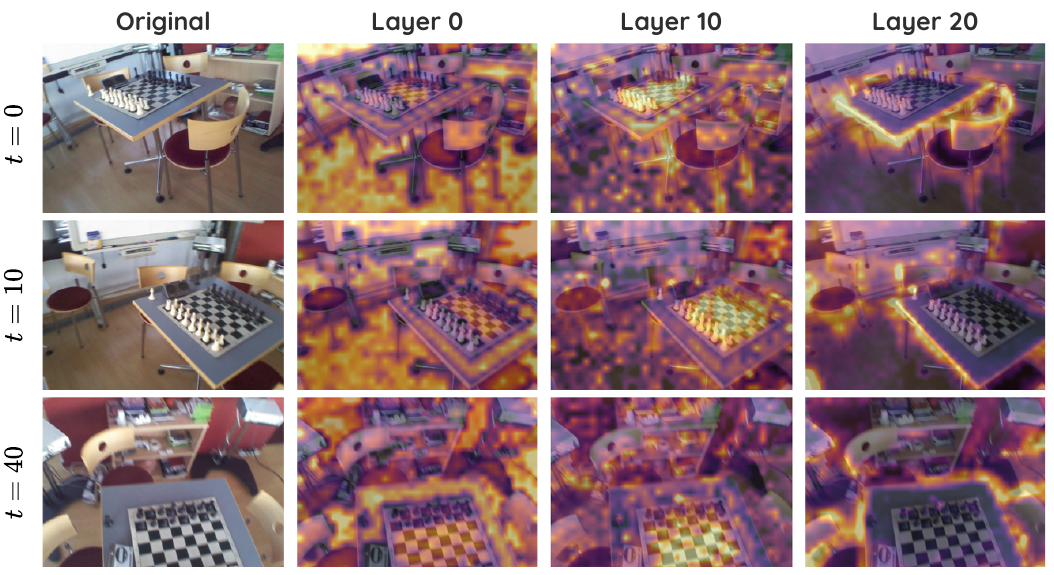}
  \vspace{-6mm}
  \caption{\textbf{Per-token FFN activation scores} across layers, progressing from high-frequency textures (shallow) to geometric structures (mid) to semantic boundaries (deep).}
  \label{fig:ffnvalue}
\end{figure}

Maintaining a fixed-size cache requires deciding \emph{which} tokens to retain.
Attention-map-based selection is the most intuitive criterion, yet modern pipelines rely on FlashAttention~\cite{dao2022flashattention,dao2024flashattention2}, which avoids materializing the full attention matrix $\mathbf{A} \in \mathbb{R}^{N \times N}$ and thus cannot expose per-token attention weights without sacrificing efficiency.
Token compression techniques developed for large models~\cite{feng2024adakv,choi2025repshift} offer a natural starting point for cache management, yet the representations in geometric transformers carry fundamentally different semantics: whereas LLM compression targets linguistically salient text tokens, geometric patch tokens undergo spatially structured nonlinear transformations that progressively encode texture, geometry, and semantic boundary information across layers (\cref{fig:ffnvalue}).
Our probing experiments empirically confirm that FFN-residual-based scoring consistently outperforms attention-weight-based, query-key-dot-product-based, and random eviction strategies across all reconstruction metrics, validating its effectiveness as a geometric saliency proxy.
Building on this insight, we propose a self-selective strategy that derives importance entirely from quantities already computed in the forward pass, requiring no additional modules.

% - - - - - - - - - - - - - - - - - - - - - - - - - - - - - - - - - - - - - - -
\noindent\textbf{\\Activation Value Rating.}
\label{sec:avr}
% - - - - - - - - - - - - - - - - - - - - - - - - - - - - - - - - - - - - - - -
While FlashAttention precludes access to attention weights, the feed-forward network (FFN) within each Transformer block operates independently of the attention kernel.
In the Pre-LN formulation,
\begin{align}
  \mathbf{h}^{(l)}
    &= \mathbf{x}^{(l-1)}
       + \mathrm{MHA}\!\bigl(\mathrm{LN}(\mathbf{x}^{(l-1)})\bigr), \\
  \mathbf{x}^{(l)}
    &= \mathbf{h}^{(l)}
       + \lambda_2^{(l)} \cdot
         \mathrm{FFN}\!\bigl(\mathrm{LN}(\mathbf{h}^{(l)})\bigr),
\end{align}
where $\lambda_2^{(l)}$ is the LayerScale~\cite{touvron2021layerscale} coefficient.
The FFN residual $\lambda_2^{(l)}\!\cdot\!\mathrm{FFN}(\mathrm{LN}(\mathbf{h}^{(l)}))$ applies a token-wise nonlinear transformation that amplifies salient representations; its magnitude naturally reflects how strongly each token is activated.
We therefore define the activation score of the $i$-th token at layer $l$ as
\begin{equation}
  s_i^{(l)}
  = \bigl\|\,\lambda_2^{(l)} \cdot
      \mathrm{FFN}\!\bigl(\mathrm{LN}(\mathbf{h}_i^{(l)})\bigr)
    \,\bigr\|_2,
\end{equation}
quantifying the representational shift induced by the FFN.
As visualized in \cref{fig:ffnvalue}, shallow layers yield high scores in textured regions, intermediate layers highlight geometrically informative patches (\eg, the checkerboard), and deep layers concentrate on semantic object boundaries, reflecting the coarse-to-fine hierarchy of Transformers.
Crucially, since the FFN residual is already computed during the forward pass, this scoring introduces zero additional memory or computation overhead and remains fully compatible with FlashAttention.

\begin{figure}[t]
  \centering
  \includegraphics[width=\columnwidth]{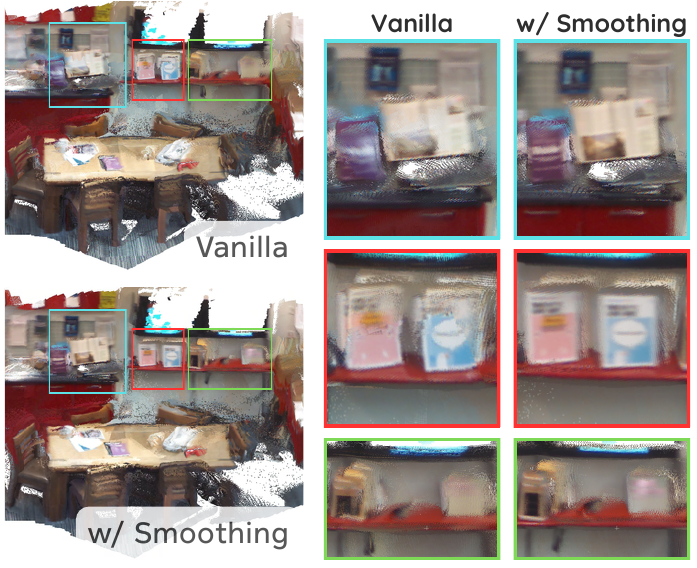}
  \vspace{-7mm}
  \caption{\textbf{Activation smoothing} effectively improves reconstruction quality over vanilla token retention.}
  \label{fig:smoothing}
\end{figure}

\textbf{\textit{Activation Smoothing.}}
Directly selecting tokens by raw activation scores tends to produce spatially fragmented retention patterns, introducing discontinuous references that degrade reconstruction sharpness (\cref{fig:smoothing}, ``Vanilla'').
A key distinction from LLM token compression is that geometric patch tokens possess inherent 2D spatial structure: neighboring patches observe overlapping scene regions and share local geometric context.
Disrupting this spatial coherence scatters the retained references across the image and destroys the local continuity that depth and point cloud heads rely upon.
We therefore apply Gaussian smoothing to the 2D activation map $\mathbf{S} \in \mathbb{R}^{H_p \times W_p}$:
\begin{equation}
  \tilde{\mathbf{S}}
  = \alpha \cdot (\mathbf{G} * \mathbf{S})
    + (1 - \alpha) \cdot \mathbf{S},
\end{equation}
where $\mathbf{G}$ is a Gaussian kernel and $\alpha$ controls the smoothing strength.
This encourages spatially coherent token groups to be retained, preserving the local context critical for accurate depth and point cloud prediction (\cref{fig:smoothing}, ``w/ Smoothing'').
Aux tokens are excluded from smoothing and retain original scores.

% - - - - - - - - - - - - - - - - - - - - - - - - - - - - - - - - - - - - - - -
\noindent\textbf{\\KV Cache Compression.}
\label{sec:kvcc}
% - - - - - - - - - - - - - - - - - - - - - - - - - - - - - - - - - - - - - - -
Given per-token activation scores, we compress each layer's cache to its budget $B^{(l)}$ (with $B = \sum_{l=0}^{L-1} B^{(l)}$).
For brevity, we omit the layer index; all operations execute independently per layer.
We partition the cache into a \emph{protected} set $\mathcal{P}$ (exempt from eviction; see \cref{sec:dap}) and an \emph{evictable} set $\mathcal{U} = \mathcal{U}_{\mathrm{hist}} \cup \mathcal{U}_{\mathrm{new}}$, with $|\mathcal{P}| + |\mathcal{U}| = |\mathcal{C}_t|$.

\textbf{\textit{Hybrid Scoring.}}
Since only current-frame tokens pass through the FFN, historical tokens lack up-to-date activation scores; we therefore adopt a dual-metric scheme.
Following~\cite{yuan2026infinitevggt}, each historical token in $\mathcal{U}_{\mathrm{hist}}$ is assigned a \emph{diversity score} $d_i = 1 - \cos(\mathbf{k}_i, \bar{\mathbf{k}})$, where $\bar{\mathbf{k}}$ denotes the centroid key vector; tokens with higher deviation from the centroid are considered more informative.
Current-frame tokens in $\mathcal{U}_{\mathrm{new}}$ directly use the activation score $s_i^{(l)}$.
After independent min-max normalization ($\hat{d}_i, \hat{s}_i \in [0,1]$), a priority coefficient $\beta \in [0,1]$ balances the two sources:
\begin{equation}
  r_i =
  \begin{cases}
    (1 - \beta) \cdot \hat{d}_i, & i \in \mathcal{U}_{\mathrm{hist}}, \\[3pt]
    \beta \cdot \hat{s}_i,        & i \in \mathcal{U}_{\mathrm{new}}.
  \end{cases}
\end{equation}
A larger $\beta$ favors current-frame tokens; a smaller $\beta$ prioritizes historical diversity.
We retain the $B^{(l)} - |\mathcal{P}|$ highest-scoring tokens from $\mathcal{U}$ and evict the rest:
\begin{equation}
  \tilde{\mathcal{C}}_t
  = \mathcal{P} \cup
    \mathrm{Top\text{-}k}\!\bigl(
      \mathcal{U},\;
      B^{(l)} - |\mathcal{P}|
    \bigr).
\end{equation}
Per-layer budgets $B^{(l)}$ are allocated proportionally to each layer's token diversity~\cite{yuan2026infinitevggt}, granting more capacity to high-diversity layers and compressing homogeneous layers more aggressively.

% -----------------------------------------------------------------------------
\subsection{Dynamic Anchor Protection}
\label{sec:dap}
% -----------------------------------------------------------------------------

Cache compression alone cannot guarantee geometric consistency: as the token composition churns through repeated eviction cycles, depth and point cloud predictions may drift when the camera moves far from previously observed regions.
This challenge is unique to geometric streaming and has no counterpart in LLM inference, where token sequences lack a shared spatial coordinate system.
We therefore introduce Dynamic Anchor Protection (DAP) to explicitly shield a small set of geometrically critical tokens $\mathcal{P} = \mathcal{P}_{\mathrm{init}} \cup \mathcal{P}_{\mathrm{hist}}$ from eviction.

\noindent\textbf{\\Global Initial Anchor.}
\label{sec:gia}
The first frame defines the world-coordinate origin for all subsequent predictions.
We permanently protect all $M$ first-frame tokens as $\mathcal{P}_{\mathrm{init}}$ to preserve coordinate-system consistency throughout inference.

\noindent\textbf{\\Historical Anchors.}
\label{sec:hca}
As the camera traverses the scene, the first frame may share no visual overlap with the current view, rendering $\mathcal{P}_{\mathrm{init}}$ alone insufficient.
We therefore adaptively register historical anchors to supply long-range geometric references.
Concretely, let $a$ be the most recent anchor frame with 3D points $\mathbf{P}_a$ and pose $\mathbf{T}_a \in SE(3)$.
We project $\mathbf{P}_a$ into the current view via $\mathbf{T}_t^{-1}\mathbf{T}_a$ and compute the coverage ratio $\rho_t$ as the fraction of points falling within the image bounds; a new anchor is registered at frame $t$ whenever $\rho_t < \tau$ and at least 100 frames have elapsed since the last registration, preventing excessive switching under rapid camera motion.
For each anchor frame $j$, we rank its patch tokens by the per-point confidence $\mathbf{c}_j \in \mathbb{R}^{N_p}$ from the point cloud head and protect only the top-$\eta$ percentile as $\mathcal{P}_{\mathrm{hist},j}$.
To prevent unbounded accumulation, the number of active anchors is capped at $K_{\max}$ with a first-in-first-out (FIFO) policy: when the limit is reached, the oldest anchor is demoted back to $\mathcal{U}_{\mathrm{hist}}$.
We adopt FIFO over more complex replacement strategies as it introduces zero additional computation and, given that older anchors are naturally more likely to have been superseded by newer, spatially closer references, empirically performs on par with alternatives.
The resulting budget overhead is bounded by
\begin{equation}
  |\mathcal{P}| \leq M + K_{\max} \cdot \lceil \eta \cdot N_p \rceil,
\end{equation}
where the first term accounts for the global initial anchor tokens and the second for all active historical anchors.
In practice, we set $K_{\max}$ and $\eta$ to small values so that the total anchor overhead remains a modest fraction of each layer's budget $B^{(l)}$, leaving sufficient capacity in the evictable pool to maintain token diversity.
SSC and DAP are complementary: the former identifies which tokens carry geometric value, while the latter guarantees that coordinate-critical references survive across eviction cycles.
Their joint design reflects the spatial and metric structure inherent to visual geometry, and we empirically validate each component in Sec.~\ref{sec:ablation} and the supplementary material.

%% file: sec/4_exp.tex
% !TEX root = ../main.tex

% \newpage

\begin{figure*}[!htbp]
  \centering
  \includegraphics[width=\textwidth]{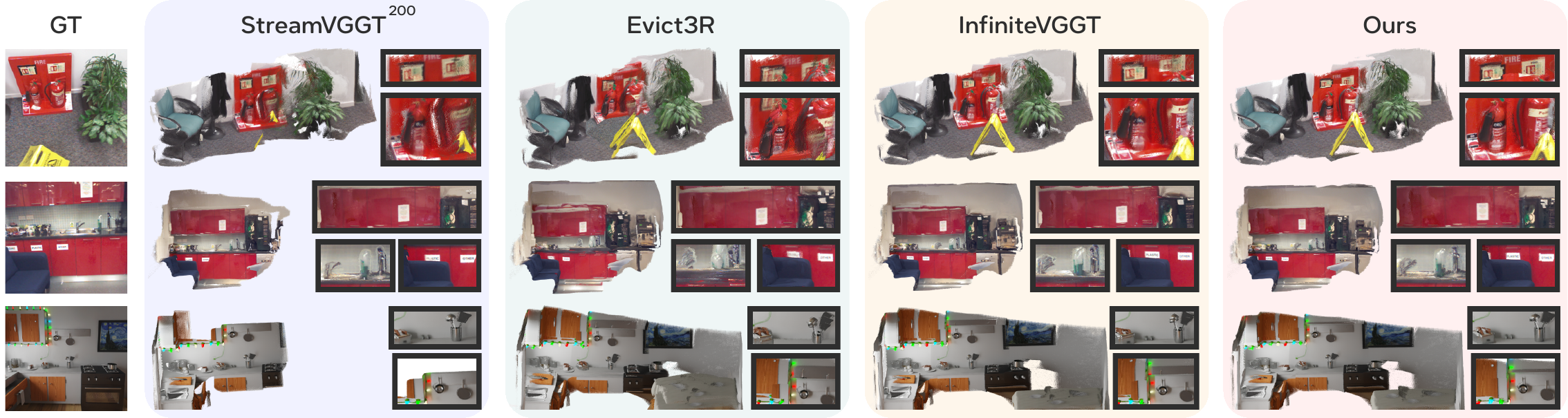}
  \caption{\textbf{Qualitative comparison on indoor scene reconstruction} (sequence length $= 500$). Each row shows a different scene with close-up insets. Note that StreamVGGT is limited to a maximum of 200 input frames due to memory constraints.}
  \label{fig:qualitative}
\end{figure*}

\input{table/quant_7s_nrgbd}

\section{Experiments}
\label{sec:experiments}

To comprehensively validate \model{} against existing streaming models, we evaluate 3D reconstruction performance across three diverse scene categories: indoor, outdoor, and ultra-long sequences (\cref{sec:exp:outdoor_ultralong}).
We further provide detailed comparisons on video depth estimation (\cref{sec:exp:depth}) and inference efficiency analysis (\cref{sec:exp:efficiency}) against causal visual geometry models.

\noindent\textbf{Baselines.}
The works most closely related to ours are Evict3R~\cite{mahdi2025evict3r} and InfiniteVGGT~\cite{yuan2026infinitevggt}, both built on the causal architecture of StreamVGGT~\cite{zheng2025streamvggt}, which serves as the full-cache reference.
InfiniteVGGT uses its default fixed budget; for Evict3R, which specifies a retention ratio rather than an absolute budget, we report both the original and a budget-matched variant (Evict3R$^\dagger$) dynamically calibrated to match \model{}.
We further compare against Spann3R~\cite{wang2024spann3r}, CUT3R~\cite{wang2025cut3r}, TTT3R~\cite{chen2025ttt3r}, and Point3R~\cite{wang2025point3r}.

\noindent\textbf{Implementation Details.}
The default cache budget is $B{=}200\text{K}$ tokens, occupying approximately 10\,GB and comfortably supporting arbitrarily long inference on consumer-grade GPUs; the ablation in \cref{sec:ablation:budget} confirms this as the most cost-effective operating point.
Within SSC, the smoothing coefficient is set to $\alpha{=}0.5$ and the hybrid scoring balance to $\beta{=}0.5$.
For DAP, the view-overlap threshold is $\tau{=}0.2$ with a minimum interval of 100 frames between anchor registrations, the anchor token retention percentile is $\eta{=}0.05$, and the maximum number of active historical anchors is $K_{\max}{=}3$.
All experiments are conducted on a single 32\,GB NVIDIA RTX 5090 GPU.

% -----------------------------------------------------------------------------
\subsection{3D Reconstruction}
\label{sec:exp:recon}

% -----------------------------------------------------------------------------
% ---- Indoor ----
\noindent\textbf{Indoor Benchmarks.}
\label{sec:exp:indoor}
We evaluate indoor 3D reconstruction on 7-Scenes~\cite{shotton2013scenes} and NRGBD~\cite{azinovic2022neural}, reporting Accuracy (Acc), Completeness (Comp), and Normal Consistency (NC).
Following~\cite{chen2025ttt3r,yuan2026infinitevggt}, we sample sequences of 100 to 500 frames at stride 2; as a more challenging stress test, we further reduce the stride to 1 on 7-Scenes, yielding full-sequence input that places greater demands on long-horizon stability.
As shown in \cref{tab:comparison}, \model{} achieves state-of-the-art performance under a constant resource budget, with its advantage becoming increasingly pronounced as sequence length grows.
StreamVGGT rapidly exhausts VRAM and cannot continue processing beyond short sequences.
Notably, the superior accuracy of \model{} relative to StreamVGGT indicates that retaining the entire cache does not represent an accuracy upper bound: redundant cached tokens can degrade reconstruction quality.
Despite this accuracy, \model{} maintains a clear inference cost advantage among causal-pipeline methods (\cref{sec:exp:efficiency}).
Qualitative results in \cref{fig:qualitative} further corroborate these findings: at 500 frames, competing methods exhibit geometric distortion and blurred details, whereas \model{} preserves sharp structures and coherent geometry throughout.

% ---- Outdoor & Ultra-long ----
\noindent\textbf{Outdoor and Ultra-Long Sequences.}
\label{sec:exp:outdoor_ultralong}
For outdoor and ultra-long sequence evaluation, we report results on ETH3D~\cite{schops2017multi} and Long3D~\cite{yuan2026infinitevggt}.
All methods receive the complete sequence as input without subsampling.
Notably, the ultra-long sequences in Long3D contain up to 10{,}000 consecutive frames.
For the particularly challenging ETH3D~\cite{schops2017multi} and Long3D~\cite{yuan2026infinitevggt} datasets, we additionally report results with an increased budget of $B{=}400\text{K}$ (denoted Ours$^{400}$) to accommodate the complexity of outdoor scenes and ultra-long sequence lengths, alongside the default $B{=}200\text{K}$ configuration.
This increase incurs only a modest overhead of approximately 1\,GB in allocated VRAM relative to the default budget, remaining well below the resource consumption of other causal baselines.

As shown in \cref{tab:outdoor_long}, \model{} delivers the best performance in both open outdoor scenes and ultra-long sequences, exhibiting stable reconstruction quality throughout.
These results confirm that \model{} can maintain cache effectiveness and compactness under complex scenes and extended inference horizons, actively filtering out noisy and redundant information to sustain reconstruction performance.
The metrics of StreamVGGT on outdoor scenes corroborate this observation directly: Evict3R and InfiniteVGGT, which are also built upon the same full-cache baseline, achieve comparable or inferior accuracy relative to StreamVGGT, whereas \model{} consistently surpasses it.

\input{table/quant_ETH3D}

\input{table/quant_video_depth}
% -----------------------------------------------------------------------------
\subsection{Video Depth Estimation}
\label{sec:exp:depth}
% -----------------------------------------------------------------------------
Beyond 3D reconstruction, we also evaluate \model{} on long-sequence video depth estimation.
Unlike direct 3D point evaluation, which can be affected by cumulative noise over long sequences, depth estimation better reflects per-frame local geometric accuracy.
\cref{tab:video_depth} reports depth metrics on Bonn~\cite{palazzolo2019refusion} and KITTI~\cite{geiger2013vision}, both of which contain dynamic objects.

On the indoor Bonn sequences, \model{} performs on par with the best baselines at shorter sequence lengths, yet maintains stable accuracy as sequences grow longer, whereas other methods exhibit increasing error accumulation.
On the outdoor driving scenes of KITTI, \model{} already surpasses the full-cache baseline StreamVGGT even at short input lengths.
We attribute this to the complexity of large-scale outdoor scenes, where redundant cached tokens introduce substantial noise during inference; the self-selective caching and anchor protection of \model{} can effectively filter such noise and ensure stable geometric inference.
As sequence length further increases, \model{} exhibits minimal metric fluctuation, demonstrating robust inference under the default $B{=}200\text{K}$ budget while outperforming baselines that consume considerably more resources.

\begin{figure*}[]
  \centering
  \includegraphics[width=\textwidth]{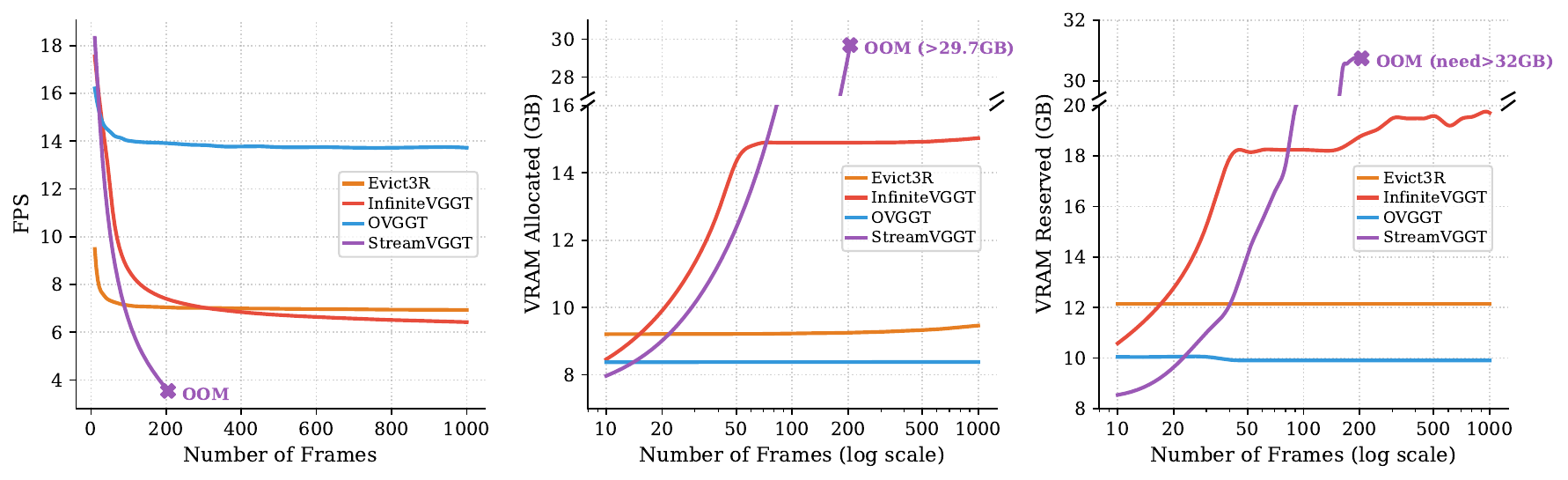}
  \caption{\textbf{Efficiency comparison.} FPS and VRAM vs.\ sequence length.}
  \label{fig:efficiency}
\end{figure*}

% -----------------------------------------------------------------------------
\subsection{Inference Efficiency}
\label{sec:exp:efficiency}
% -----------------------------------------------------------------------------

\cref{fig:efficiency} profiles FPS and peak VRAM (both allocated and reserved) against sequence length on the same 7-Scenes configuration reported in \cref{tab:comparison}, where \model{} already holds a clear accuracy advantage.
In throughput, \model{} achieves a substantial lead: Evict3R and InfiniteVGGT maintain stable but sub-real-time frame rates, while StreamVGGT's per-step cost grows with the accumulated cache, triggering OOM beyond ${\sim}200$ frames.
In VRAM, StreamVGGT exceeds 32\,GB reserved memory at that point, and InfiniteVGGT also incurs high overhead due to its large default budget.
Evict3R's allocated VRAM is only ${\sim}1$\,GB above \model{}, yet its reserved memory is considerably higher because materializing attention maps for eviction precludes FlashAttention.
By contrast, \model{} retains full FlashAttention compatibility, achieving both the lowest memory footprint and the highest throughput, realizing truly $O(1)$ constant-cost inference per frame.

% =============================================================================
\section{Ablation Studies}
\label{sec:ablation}
% =============================================================================

% -----------------------------------------------------------------------------
\noindent\textbf{Cache Budget Capacity.}
\label{sec:ablation:budget}
\cref{tab:ablation_budget} reports reconstruction metrics under varying cache budgets.
Performance degrades noticeably with an excessively small budget, stabilizes at $200\text{K}$, and yields diminishing returns beyond this point.
We therefore adopt $B{=}200\text{K}$ as the default, which provides sufficient accuracy across typical scenes while fitting within a 12\,GB VRAM envelope, enabling deployment on standard consumer-grade GPUs.

\input{table/ablation_budget}

% -----------------------------------------------------------------------------
\noindent\textbf{\\Effect of Activation Smoothing.}
\label{sec:ablation:smoothing}
\cref{tab:ablation_smoothing} reports the effect of smoothing coefficient $\alpha$ (mean metrics).
Increasing $\alpha$ progressively improves accuracy by encouraging spatially coherent token retention, but excessively high values introduce over-averaging artifacts in point cloud visualizations.
We set $\alpha{=}0.5$ as a balanced default that preserves reconstruction sharpness while maintaining stable reference retention.

% -----------------------------------------------------------------------------
\noindent\textbf{\\Hybrid Scoring Balance.}
\label{sec:ablation:hybrid}
The coefficient $\beta$ balances current-frame activation scores against historical key-vector diversity.
As shown in \cref{tab:ablation_hybrid}, which reports median reconstruction metrics on 7-Scenes, over-relying on either source degrades quality: small $\beta$ retains excessively scattered features, while large $\beta$ neglects spatial coverage.
The optimum at $\beta{=}0.5$ confirms the necessity of combining both criteria.

\input{table/ablation_smooth_score}

\input{table/ablation_anchor}

% -----------------------------------------------------------------------------
\noindent\textbf{\\Anchor Impact on Long-Range Stability.}
\label{sec:ablation:anchor}
To isolate DAP's contribution, we evaluate three configurations on KITTI depth estimation at 500 frames, stratified into near ($15$\text{--}$35$\,units) and far ($>35$\,units) ranges:
no anchoring, Global Initial Anchor only ($\mathcal{P}_{\mathrm{init}}$), and full DAP ($\mathcal{P}_{\mathrm{init}} \cup \mathcal{P}_{\mathrm{hist}}$).
As shown in \cref{tab:ablation_anchors}, the Global Initial Anchor alone yields substantial gains, and adding Historical Anchors nearly doubles the improvement, confirming that both mechanisms are essential and complementary for suppressing long-range geometric drift.

%% file: table/quant_7s_nrgbd.tex
% Requires: \usepackage{booktabs, xcolor, graphicx, multirow}

\begin{table*}[!htbp]
\newcommand{\hc}[2]{{\setlength{\fboxsep}{1pt}\colorbox{#1}{#2}}}
\caption{Quantitative comparison on the 7-Scenes~\cite{shotton2013scenes} and NRGBD~\cite{azinovic2022neural} datasets across different sequence lengths. Evict3R$^\dagger$ denotes Evict3R with its pruning rate dynamically calibrated to match our constant budget. The \hc{red!20}{best} and \hc{orange!20}{second-best} results are highlighted.
}
\label{tab:comparison}
\centering
\renewcommand{\arraystretch}{0.9}
\setlength{\tabcolsep}{2pt}
\resizebox{\textwidth}{!}{%
\begin{tabular}{l c cc cc cc c cc cc cc}
\toprule
\multirow{3}{*}[-6pt]{Method} & \multicolumn{7}{c}{7-Scenes} & \multicolumn{7}{c}{NRGBD} \\
\cmidrule(lr){2-8} \cmidrule(lr){9-15}
 & \multirow{2}{*}[-3pt]{\begin{tabular}[c]{@{}c@{}}Seq.\\[-2pt] Len.\end{tabular}}
 & \multicolumn{2}{c}{Acc $\downarrow$} & \multicolumn{2}{c}{Comp $\downarrow$} & \multicolumn{2}{c}{NC $\uparrow$}
 & \multirow{2}{*}[-3pt]{\begin{tabular}[c]{@{}c@{}}Seq.\\[-2pt] Len.\end{tabular}}
 & \multicolumn{2}{c}{Acc $\downarrow$} & \multicolumn{2}{c}{Comp $\downarrow$} & \multicolumn{2}{c}{NC $\uparrow$} \\
\cmidrule(lr){3-4} \cmidrule(lr){5-6} \cmidrule(lr){7-8}
\cmidrule(lr){10-11} \cmidrule(lr){12-13} \cmidrule(lr){14-15}
 &  & Mean & Med. & Mean & Med. & Mean & Med.
 &  & Mean & Med. & Mean & Med. & Mean & Med. \\
\midrule
%% ===== Sequence Length: 200 / 100 =====
Spann3R~\cite{wang2024spann3r}          & \colorbox{teal!5}{200\phantom{0}} & $0.215$ & $0.131$ & $0.122$ & $0.063$ & $0.535$ & $0.550$
                 & \colorbox{teal!5}{100\phantom{0}} & $0.111$ & $0.069$ & $0.045$ & $0.015$ & $0.636$ & $0.733$ \\
CUT3R~\cite{wang2025cut3r}             & \colorbox{teal!5}{\phantom{0000}}  & $0.087$ & $0.048$ & $0.045$ & $0.014$ & $0.566$ & $0.601$
                 & \colorbox{teal!5}{\phantom{0000}} & $0.039$ & $0.024$ & $0.013$ & $0.004$ & $0.645$ & $0.748$ \\
Point3R~\cite{wang2025point3r}           & \colorbox{teal!5}{\phantom{0000}} & $0.041$ & $0.019$ & \hc{orange!20}{$0.023$} & \hc{orange!20}{$0.006$} & $0.579$ & $0.622$
                 & \colorbox{teal!5}{\phantom{0000}} & $0.046$ & $0.028$ & $0.016$ & $0.004$ & $0.662$ & $0.775$ \\
TTT3R~\cite{chen2025ttt3r}             & \colorbox{teal!5}{\phantom{0000}} & \hc{orange!20}{$0.027$} & $0.015$ & \hc{orange!20}{$0.023$} & \hc{red!20}{$0.005$} & $0.582$ & $0.627$
                 & \colorbox{teal!5}{\phantom{0000}} & $0.031$ & $0.019$ & \hc{red!20}{$0.012$} & $0.004$ & $0.650$ & $0.756$ \\
StreamVGGT~\cite{zheng2025streamvggt}        & \colorbox{teal!5}{\phantom{0000}} & $0.038$ & $0.014$ & $0.029$ & $0.007$ & $0.583$ & $0.628$
                 & \colorbox{teal!5}{\phantom{0000}} & \hc{orange!20}{$0.024$} & \hc{red!20}{$0.014$} & $0.013$ & \hc{red!20}{$0.003$} & $0.663$ & $0.777$ \\
Evict3R~\cite{mahdi2025evict3r}          & \colorbox{teal!5}{\phantom{0000}} & \multicolumn{6}{c}{\smash{\hc{gray!12}{\makebox[17em]{\textcolor{gray!60}{\textit{OOM}}}}}}
                 & \colorbox{teal!5}{\phantom{0000}} & $0.025$ & \hc{orange!20}{$0.015$} & $0.013$ & \hc{red!20}{$0.003$} & $0.664$ & $0.781$ \\
Evict3R$^\dagger$~\cite{mahdi2025evict3r}          & \colorbox{teal!5}{\phantom{0000}} & $0.037$ & \hc{orange!20}{$0.013$} & $0.027$ & $0.007$ & \hc{orange!20}{$0.584$} & \hc{orange!20}{$0.631$}
                 & \colorbox{teal!5}{\phantom{0000}} & $0.031$ & $0.020$ & $0.013$ & \hc{red!20}{$0.003$} & $0.665$ & \hc{orange!20}{$0.791$} \\
InfiniteVGGT~\cite{yuan2026infinitevggt}      & \colorbox{teal!5}{\phantom{0000}} & $0.046$ & $0.016$ & $0.031$ & $0.008$ & $0.582$ & $0.627$
                 & \colorbox{teal!5}{\phantom{0000}} & $0.035$ & $0.022$ & $0.014$ & \hc{red!20}{$0.003$} & \hc{orange!20}{$0.669$} & $0.787$ \\
\textbf{Ours}    & \colorbox{teal!5}{\phantom{0000}} & \hc{red!20}{$0.024$} & \hc{red!20}{$0.008$} & \hc{red!20}{$0.021$} & \hc{red!20}{$0.005$} & \hc{red!20}{$0.587$} & \hc{red!20}{$0.635$}
                 & \colorbox{teal!5}{\phantom{0000}} & \hc{red!20}{$0.022$} & \hc{red!20}{$0.014$} & \hc{red!20}{$0.012$} & \hc{red!20}{$0.003$} & \hc{red!20}{$0.672$} & \hc{red!20}{$0.796$} \\
\midrule
%% ===== Sequence Length: 500 / 300 =====
Spann3R~\cite{wang2024spann3r}          & \colorbox{teal!5}{500\phantom{0}} & $0.343$ & $0.263$ & $0.154$ & $0.085$ & $0.515$ & $0.521$
                 & \colorbox{teal!5}{300\phantom{0}} & $0.346$ & $0.221$ & $0.175$ & $0.099$ & $0.558$ & $0.586$ \\
CUT3R~\cite{wang2025cut3r}            & \colorbox{teal!5}{\phantom{0000}} & $0.194$ & $0.143$ & $0.092$ & $0.034$ & $0.527$ & $0.538$
                 & \colorbox{teal!5}{\phantom{0000}} & $0.244$ & $0.136$ & $0.081$ & $0.019$ & $0.575$ & $0.613$ \\
Point3R~\cite{wang2025point3r}           & \colorbox{teal!5}{\phantom{0000}} & $0.056$ & $0.025$ & $0.031$ & $0.012$ & $0.555$ & $0.584$
                 & \colorbox{teal!5}{\phantom{0000}} & $0.076$ & $0.042$ & \hc{red!20}{$0.014$} & \hc{orange!20}{$0.004$} & $0.624$ & $0.707$ \\
TTT3R~\cite{chen2025ttt3r}            & \colorbox{teal!5}{\phantom{0000}} & $0.065$ & $0.037$ & $0.030$ & $0.006$ & $0.552$ & $0.578$
                 & \colorbox{teal!5}{\phantom{0000}} & $0.102$ & $0.043$ & $0.026$ & $0.005$ & $0.610$ & $0.678$ \\
StreamVGGT~\cite{zheng2025streamvggt}        & \colorbox{teal!5}{\phantom{0000}} & \multicolumn{6}{c}{\smash{\hc{gray!12}{\makebox[17em]{\textcolor{gray!60}{\textit{OOM}}}}}}
                 & \colorbox{teal!5}{\phantom{0000}} & \multicolumn{6}{c}{\smash{\hc{gray!12}{\makebox[17em]{\textcolor{gray!60}{\textit{OOM}}}}}} \\
Evict3R~\cite{mahdi2025evict3r}          & \colorbox{teal!5}{\phantom{0000}} & \multicolumn{6}{c}{\smash{\hc{gray!12}{\makebox[17em]{\textcolor{gray!60}{\textit{OOM}}}}}}
                 & \colorbox{teal!5}{\phantom{0000}} & \multicolumn{6}{c}{\smash{\hc{gray!12}{\makebox[17em]{\textcolor{gray!60}{\textit{OOM}}}}}} \\
Evict3R$^\dagger$~\cite{mahdi2025evict3r}          & \colorbox{teal!5}{\phantom{0000}} & $0.042$ & $0.016$ & $0.026$ & \hc{orange!20}{$0.005$} & \hc{orange!20}{$0.559$} & \hc{orange!20}{$0.589$} 
                 & \colorbox{teal!5}{\phantom{0000}} & \hc{orange!20}{$0.042$} & \hc{orange!20}{$0.026$} & $0.017$ &   \hc{orange!20}{$0.004$} & $0.640$ & $0.739$  \\
InfiniteVGGT~\cite{yuan2026infinitevggt}     & \colorbox{teal!5}{\phantom{0000}} & \hc{orange!20}{$0.040$} & \hc{orange!20}{$0.015$} & \hc{orange!20}{$0.024$} & \hc{orange!20}{$0.005$} & \hc{red!20}{$0.561$} & \hc{red!20}{$0.593$}
                 & \colorbox{teal!5}{\phantom{0000}} & $0.053$ & $0.031$ & $0.024$ & $0.005$ & \hc{red!20}{$0.646$} & \hc{red!20}{$0.751$} \\
\textbf{Ours}    & \colorbox{teal!5}{\phantom{0000}} & \hc{red!20}{$0.031$} & \hc{red!20}{$0.011$} & \hc{red!20}{$0.020$} & \hc{red!20}{$0.003$} & \hc{red!20}{$0.561$} & \hc{red!20}{$0.593$}
                 & \colorbox{teal!5}{\phantom{0000}} & \hc{red!20}{$0.037$} & \hc{red!20}{$0.022$} & \hc{orange!20}{$0.015$} & \hc{red!20}{$0.003$} & \hc{orange!20}{$0.642$} & \hc{orange!20}{$0.740$} \\
\midrule
%% ===== Sequence Length: 1000 / 500 =====
Spann3R~\cite{wang2024spann3r}          & \colorbox{teal!5}{1000} & $0.340$ & $0.262$ & $0.154$ & $0.092$ & $0.508$ & $0.510$
                 & \colorbox{teal!5}{500\phantom{0}} & $0.516$ & $0.342$ & $0.225$ & $0.130$ & $0.552$ & $0.578$ \\
CUT3R~\cite{wang2025cut3r}            & \colorbox{teal!5}{\phantom{0000}} & $0.240$ & $0.166$ & $0.102$ & $0.015$ & $0.513$ & $0.516$
                 & \colorbox{teal!5}{\phantom{0000}} & $0.328$ & $0.247$ & $0.157$ & $0.085$ & $0.562$ & $0.592$ \\
Point3R~\cite{wang2025point3r}          & \colorbox{teal!5}{\phantom{0000}} & $0.068$ & \hc{orange!20}{$0.028$} & \hc{orange!20}{$0.025$} & \hc{orange!20}{$0.006$} & \hc{orange!20}{$0.533$} & \hc{orange!20}{$0.549$}
                 & \colorbox{teal!5}{\phantom{0000}} & $0.116$ & $0.049$ & \hc{orange!20}{$0.027$} & \hc{red!20}{$0.004$} & $0.620$ & $0.698$ \\
TTT3R~\cite{chen2025ttt3r}            & \colorbox{teal!5}{\phantom{0000}} & $0.126$ & $0.080$ & $0.050$ & $0.010$ & $0.525$ & $0.535$
                 & \colorbox{teal!5}{\phantom{0000}} & $0.169$ & $0.082$ & $0.096$ & $0.015$ & $0.594$ & $0.647$ \\
StreamVGGT~\cite{zheng2025streamvggt}       & \colorbox{teal!5}{\phantom{0000}} & \multicolumn{6}{c}{\smash{\hc{gray!12}{\makebox[17em]{\textcolor{gray!60}{\textit{OOM}}}}}}
                 & \colorbox{teal!5}{\phantom{0000}} & \multicolumn{6}{c}{\smash{\hc{gray!12}{\makebox[17em]{\textcolor{gray!60}{\textit{OOM}}}}}} \\
Evict3R~\cite{mahdi2025evict3r}         & \colorbox{teal!5}{\phantom{0000}} & \multicolumn{6}{c}{\smash{\hc{gray!12}{\makebox[17em]{\textcolor{gray!60}{\textit{OOM}}}}}}
                 & \colorbox{teal!5}{\phantom{0000}} & \multicolumn{6}{c}{\smash{\hc{gray!12}{\makebox[17em]{\textcolor{gray!60}{\textit{OOM}}}}}} \\
Evict3R$^\dagger$~\cite{mahdi2025evict3r}         & \colorbox{teal!5}{\phantom{0000}} & $0.134$ & $0.059$ & $0.052$ & $0.009$ & $0.531$ & $0.545$
                 & \colorbox{teal!5}{\phantom{0000}} & $0.072$ & \hc{orange!20}{$0.040$} & \hc{red!20}{$0.026$} & \hc{orange!20}{$0.006$} & \hc{orange!20}{$0.641$} & \hc{orange!20}{$0.739$} \\
InfiniteVGGT~\cite{yuan2026infinitevggt}     & \colorbox{teal!5}{\phantom{0000}} & \hc{orange!20}{$0.061$} & $0.031$ & $0.035$ & $0.014$ & \hc{red!20}{$0.537$} & \hc{red!20}{$0.554$}
                 & \colorbox{teal!5}{\phantom{0000}} & \hc{orange!20}{$0.070$} & $0.046$ & $0.037$ & $0.008$ & \hc{red!20}{$0.642$} & \hc{red!20}{$0.743$} \\
\textbf{Ours}    & \colorbox{teal!5}{\phantom{0000}} & \hc{red!20}{$0.039$} & \hc{red!20}{$0.014$} & \hc{red!20}{$0.020$} & \hc{red!20}{$0.003$} & \hc{red!20}{$0.537$} & \hc{red!20}{$0.554$}
                 & \colorbox{teal!5}{\phantom{0000}} & \hc{red!20}{$0.054$} & \hc{red!20}{$0.032$} & \hc{red!20}{$0.026$} & \hc{orange!20}{$0.006$} & $0.637$ & $0.732$ \\
\bottomrule
\end{tabular}%
}
\end{table*}

%% file: table/quant_ETH3D.tex
% Requires: \usepackage{booktabs, xcolor, graphicx, multirow}

\begin{table*}[!htbp]
\newcommand{\hc}[2]{{\setlength{\fboxsep}{1pt}\colorbox{#1}{#2}}}
\caption{Quantitative comparison on full sequences of ETH3D~\cite{schops2017multi} and Long3D~\cite{yuan2026infinitevggt} datasets. Ours$^{200}$ and Ours$^{400}$ denote results with the default 200K and an increased 400K token budget, respectively. \hc{red!20}{Best} and \hc{orange!20}{second best} results highlighted.}
\label{tab:outdoor_long}
\centering
\renewcommand{\arraystretch}{0.9}
\setlength{\tabcolsep}{4pt}
\resizebox{\textwidth}{!}{%
\begin{tabular}{l cc cc cc cc cc cc}
\toprule
\multirow{3}{*}[-3pt]{Method} & \multicolumn{6}{c}{ETH3D (Outdoor)} & \multicolumn{6}{c}{Long3D (Ultra-Long)} \\
\cmidrule(lr){2-7} \cmidrule(lr){8-13}
 & \multicolumn{2}{c}{Acc $\downarrow$} & \multicolumn{2}{c}{Comp $\downarrow$} & \multicolumn{2}{c}{NC $\uparrow$}
 & \multicolumn{2}{c}{Acc $\downarrow$} & \multicolumn{2}{c}{Comp $\downarrow$} & \multicolumn{2}{c}{NC $\uparrow$} \\
\cmidrule(lr){2-3} \cmidrule(lr){4-5} \cmidrule(lr){6-7}
\cmidrule(lr){8-9} \cmidrule(lr){10-11} \cmidrule(lr){12-13}
 & Mean & Med. & Mean & Med. & Mean & Med. 
 & Mean & Med. & Mean & Med. & Mean & Med. \\
\midrule
CUT3R~\cite{wang2025cut3r}        & $0.940$ & $0.607$ & $0.709$ & $0.374$ & $0.718$ & $0.812$ 
             & $6.189$ & $2.405$ & $1.921$ & $1.535$ & $0.501$ & $0.497$ \\
TTT3R~\cite{chen2025ttt3r}        & \hc{orange!20}{$0.598$} & $0.374$ & $0.585$ & $0.223$ & $0.728$ & $0.826$ 
             & $7.341$ & $2.018$ & $1.455$ & $1.235$ & \hc{orange!20}{$0.509$} & $0.503$ \\
StreamVGGT~\cite{zheng2025streamvggt}   & $0.601$ & \hc{orange!20}{$0.369$} & $0.442$ & $0.169$ & $0.791$ & $0.933$ 
             & \multicolumn{6}{c}{\smash{\hc{gray!12}{\makebox[19em]{\textcolor{gray!60}{\textit{OOM}}}}}} \\
Evict3R$^\dagger$~\cite{mahdi2025evict3r}      & $0.605$ & $0.375$ & $0.442$ & $0.163$ & \hc{orange!20}{$0.792$} & \hc{red!20}{$0.934$} 
             & $4.928$ & $2.710$ & $0.715$ & $0.204$ & $0.507$ & $0.504$ \\
InfiniteVGGT~\cite{yuan2026infinitevggt} & $0.603$ & $0.371$ & $0.444$ & $0.169$ & \hc{orange!20}{$0.792$} & $0.933$ 
             & $4.344$ & $3.668$ & $0.974$ & $0.205$ & \hc{red!20}{$0.517$} & \hc{red!20}{$0.525$} \\
\textbf{Ours}$^{200}$ & $0.628$ & $0.396$ & \hc{red!20}{$0.380$} & \hc{orange!20}{$0.121$} & $0.790$ & \hc{red!20}{$0.934$} 
             & \hc{orange!20}{$2.453$} & \hc{orange!20}{$1.794$} & \hc{red!20}{$0.390$} & \hc{red!20}{$0.060$} & $0.507$ & \hc{orange!20}{$0.509$} \\
\textbf{Ours}$^{400}$ & \hc{red!20}{$0.535$} & \hc{red!20}{$0.317$} & \hc{orange!20}{$0.394$} & \hc{red!20}{$0.107$} & \hc{red!20}{$0.793$} & \hc{red!20}{$0.934$} 
             & \hc{red!20}{$2.449$} & \hc{red!20}{$1.675$} & \hc{orange!20}{$0.542$} & \hc{orange!20}{$0.151$} & $0.507$ & \hc{orange!20}{$0.509$} \\ 
\bottomrule
\end{tabular}%
}
\end{table*}

%% file: table/quant_video_depth.tex
\begin{table*}[!htbp]
    \newcommand{\hc}[2]{{\setlength{\fboxsep}{1pt}\colorbox{#1}{#2}}}
    \centering
    \caption{Video depth evaluation on Bonn~\cite{palazzolo2019refusion} and KITTI~\cite{geiger2013vision} across different sequence lengths. \hc{red!20}{Best} results highlighted.}
    \label{tab:video_depth}
    \renewcommand{\arraystretch}{0.9}
    \setlength{\tabcolsep}{4pt}
    \resizebox{\linewidth}{!}{
        \begin{tabular}{@{}l c@{\hskip 8pt}c@{\hskip 8pt}c @{\hskip 8pt} c@{\hskip 8pt}c@{\hskip 8pt}c @{\hskip 12pt} c@{\hskip 8pt}c@{\hskip 8pt}c @{\hskip 8pt} c@{\hskip 8pt}c@{\hskip 8pt}c@{}}
            \toprule
            & \multicolumn{6}{c}{Bonn} & \multicolumn{6}{c}{KITTI} \\
            \cmidrule(lr){2-7} \cmidrule(lr){8-13}
            \multirow{2}{*}[-3pt]{Method}
            & \multicolumn{3}{c}{Abs Rel $\downarrow$} & \multicolumn{3}{c}{$\delta \!<\! 1.25$ $\uparrow$}
            & \multicolumn{3}{c}{Abs Rel $\downarrow$} & \multicolumn{3}{c}{$\delta \!<\! 1.25$ $\uparrow$} \\
            \cmidrule(lr){2-4} \cmidrule(lr){5-7} \cmidrule(lr){8-10} \cmidrule(lr){11-13}
            & 100 & 300 & 500
            & 100 & 300 & 500
            & 100 & 300 & 500
            & 100 & 300 & 500 \\
            \midrule
            StreamVGGT~\cite{zheng2025streamvggt}
                & \hc{red!20}{$0.055$} & -- & --
                & $0.974$ & -- & --
                & $0.166$ & -- & --
                & $0.740$ & -- & -- \\
            Evict3R$^\dagger$~\cite{mahdi2025evict3r}
                & $0.063$ & $0.072$ & $0.072$
                & $0.963$ & $0.951$ & $0.957$
                & $0.192$ & $0.213$ & $0.198$
                & $0.693$ & $0.700$ & $0.705$ \\
            InfiniteVGGT~\cite{yuan2026infinitevggt}
                & $0.056$ & $0.073$ & $0.070$
                & \hc{red!20}{$0.975$} & \hc{red!20}{$0.957$} & \hc{red!20}{$0.960$}
                & $0.165$ & $0.249$ & $0.257$
                & $0.742$ & $0.556$ & $0.577$ \\
            \textbf{Ours}
                & \hc{red!20}{$0.055$} & \hc{red!20}{$0.071$} & \hc{red!20}{$0.067$}
                & $0.974$ & $0.956$ & $0.959$
                & \hc{red!20}{$0.128$} & \hc{red!20}{$0.133$} & \hc{red!20}{$0.135$}
                & \hc{red!20}{$0.839$} & \hc{red!20}{$0.844$} & \hc{red!20}{$0.839$} \\
            \bottomrule
        \end{tabular}
    }
\end{table*}

%% file: table/ablation_budget.tex
\begin{table*}[!htbp]
    \newcommand{\hc}[2]{{\setlength{\fboxsep}{1pt}\colorbox{#1}{#2}}}
    \centering
    \renewcommand{\arraystretch}{0.9}
    \setlength{\tabcolsep}{3pt}
    \caption{\textbf{Effect of cache budget.} Reconstruction metrics on 7-Scenes and NRGBD at 300 frames. \hc{red!20}{Best} per column.}
    \label{tab:ablation_budget}
    \resizebox{0.97\textwidth}{!}{
        \begin{tabular}{@{}l cc cc cc c cc cc cc c@{}}
            \toprule
            \multirow{3}{*}[-2pt]{\textbf{Budget}} & \multicolumn{7}{c}{\textbf{7-Scenes}} & \multicolumn{7}{c}{\textbf{NRGBD}} \\
            \cmidrule(lr){2-8} \cmidrule(l){9-15}
            & \multicolumn{2}{c}{Acc $\downarrow$} & \multicolumn{2}{c}{Comp $\downarrow$} & \multicolumn{2}{c}{NC $\uparrow$} & \multirow{2}{*}{CD $\downarrow$} & \multicolumn{2}{c}{Acc $\downarrow$} & \multicolumn{2}{c}{Comp $\downarrow$} & \multicolumn{2}{c}{NC $\uparrow$} & \multirow{2}{*}{CD $\downarrow$} \\
            \cmidrule(lr){2-3} \cmidrule(lr){4-5} \cmidrule(lr){6-7} \cmidrule(lr){9-10} \cmidrule(lr){11-12} \cmidrule(lr){13-14}
            & Mean & Med. & Mean & Med. & Mean & Med. & & Mean & Med. & Mean & Med. & Mean & Med. & \\
            \midrule
            
            \colorbox{teal!5}{\makebox[2.5em][l]{100K}} & $0.059$ & $0.035$ & $0.030$ & $0.006$ & $0.565$ & $0.598$ & $0.043$ & \hc{red!20}{$0.037$} & $0.023$ & \hc{red!20}{$0.013$} & \hc{red!20}{$0.003$} & $0.636$ & $0.730$ & \hc{red!20}{$0.041$} \\
            
            \colorbox{teal!5}{\makebox[2.5em][l]{\textbf{200K}}} & $0.026$ & $0.009$ & \hc{red!20}{$0.020$} & \hc{red!20}{$0.004$} & $0.571$ & $0.609$ & \hc{red!20}{$0.032$} & \hc{red!20}{$0.037$} & $0.022$ & $0.015$ & \hc{red!20}{$0.003$} & \hc{red!20}{$0.642$} & {$0.740$} & $0.042$ \\
            
            \colorbox{teal!5}{\makebox[2.5em][l]{500K}} & \hc{red!20}{$0.025$} & \hc{red!20}{$0.008$} & $0.021$ & $0.005$ & $0.570$ & $0.608$ & $0.033$ & $0.038$ & \hc{red!20}{$0.021$} & $0.017$ & \hc{red!20}{$0.003$} & $0.640$ & $0.736$ & $0.043$ \\
            
            \colorbox{teal!5}{\makebox[2.5em][l]{750K}} & $0.027$ & $0.009$ & $0.022$ & $0.005$ & $0.571$ & $0.609$ & $0.035$ & $0.041$ & $0.023$ & $0.018$ & $0.004$ & $0.639$ & $0.736$ & $0.046$ \\
            
            \colorbox{teal!5}{\makebox[2.5em][l]{1M}} & $0.028$ & $0.010$ & $0.023$ & $0.005$ & \hc{red!20}{$0.572$} & \hc{red!20}{$0.610$} & $0.036$ & $0.044$ & $0.024$ & $0.019$ & $0.004$ & \hc{red!20}{$0.642$} & \hc{red!20}{$0.741$} & $0.048$ \\
            
            \bottomrule
        \end{tabular}
    }
\end{table*}

%% file: table/ablation_smooth_score.tex
\begin{table*}[!htbp]
    \newcommand{\hc}[2]{{\setlength{\fboxsep}{1pt}\colorbox{#1}{#2}}}
    \centering
    \renewcommand{\arraystretch}{0.9}
    % ==========================================
    % Left: Activation Smoothing
    % ==========================================
    \begin{minipage}[t]{0.7\linewidth}
        \caption{\textbf{Activation smoothing coefficient.} Reconstruction on 7-Scenes and NRGBD at 300 frames.}
        \label{tab:ablation_smoothing}
        \setlength{\tabcolsep}{4pt}
        \resizebox{\linewidth}{!}{
            \begin{tabular}{@{}lc cccc cccc@{}}
                \toprule
                \multirow{2}{*}{\textbf{Smooth}} & \multirow{2}{*}{$\alpha$} & \multicolumn{4}{c}{\textbf{7-Scenes}} & \multicolumn{4}{c}{\textbf{NRGBD}} \\
                \cmidrule(lr){3-6} \cmidrule(l){7-10}
                & & Acc $\downarrow$ & Comp $\downarrow$ & NC $\uparrow$ & CD $\downarrow$ & Acc $\downarrow$ & Comp $\downarrow$ & NC $\uparrow$ & CD $\downarrow$ \\
                \midrule
                
                w/o & \colorbox{teal!5}{0.0} & $0.027$ & $0.021$ & $0.571$ & $0.033$ & $0.039$ & $0.015$ & $0.643$ & $0.044$ \\
                
                \midrule
                \multirow{4}{*}{\textbf{w/}} 
                & \colorbox{teal!5}{0.1} & $0.027$ & $0.020$ & $0.571$ & $0.033$ & $0.038$ & \hc{red!20}{$0.014$} & $0.642$ & $0.044$ \\
                & \colorbox{teal!5}{0.3} & $0.028$ & $0.021$ & $0.571$ & $0.033$ & \hc{red!20}{$0.037$} & $0.015$ & $0.644$ & \hc{red!20}{$0.042$} \\
                & \colorbox{teal!5}{\textbf{0.5}} & \hc{red!20}{$0.026$} & $0.020$ & $0.571$ & \hc{red!20}{$0.032$} & \hc{red!20}{$0.037$} & $0.015$ & $0.642$ & \hc{red!20}{$0.042$} \\
                & \colorbox{teal!5}{0.9} & $0.027$ & \hc{red!20}{$0.019$} & $0.571$ & $0.033$ & $0.039$ & $0.016$ & \hc{red!20}{$0.648$} & $0.045$ \\
                \bottomrule
            \end{tabular}
        }
    \end{minipage}\hfill
    % ==========================================
    % Right: Hybrid Scoring
    % ==========================================
    \begin{minipage}[t]{0.265\linewidth}
        \caption{{Hybrid scoring balance.} 300 frames on 7-Scenes.}
        \label{tab:ablation_hybrid}
        \setlength{\tabcolsep}{4pt}
        \resizebox{\linewidth}{!}{
            \begin{tabular}{@{}l ccc@{}}
                \toprule
                $\beta$ & Acc $\downarrow$ & NC $\uparrow$ & CD $\downarrow$ \\
                \midrule
                
                \colorbox{teal!5}{\makebox[1.5em][l]{0.1}} & $0.021$ & $0.607$ & $0.024$ \\
                \colorbox{teal!5}{\makebox[1.5em][l]{0.3}} & $0.013$ & $0.606$ & $0.017$ \\
                \colorbox{teal!5}{\makebox[1.5em][l]{\textbf{0.5}}} & \hc{red!20}{$0.009$} & $0.609$ & \hc{red!20}{$0.013$} \\
                \colorbox{teal!5}{\makebox[1.5em][l]{0.7}} & $0.010$ & $0.609$ & $0.014$ \\
                \colorbox{teal!5}{\makebox[1.5em][l]{0.9}} & $0.012$ & \hc{red!20}{$0.611$} & $0.016$ \\
                \bottomrule
            \end{tabular}
        }
    \end{minipage}
    
\end{table*}

%% file: table/ablation_anchor.tex
\begin{table}[!htbp]
    \newcommand{\hc}[2]{{\setlength{\fboxsep}{1pt}\colorbox{#1}{#2}}}
    \centering
    \renewcommand{\arraystretch}{0.9}
    \setlength{\tabcolsep}{4pt}
    \caption{\textbf{Effect of Dynamic Anchor Protection.} Depth estimation improvement over the no-anchor baseline on KITTI~\cite{geiger2013vision} at 500 frames, split by depth range.}
    \label{tab:ablation_anchors}
    \resizebox{\linewidth}{!}{
        \begin{tabular}{@{}cc ccc ccc@{}}
            \toprule
            \multirow{2}{*}{\textbf{$\mathcal{P}_{\mathrm{init}}$}} & \multirow{2}{*}{\textbf{$\mathcal{P}_{\mathrm{hist}}$}} & \multicolumn{3}{c}{\textbf{Far} ($>35$\,units)} & \multicolumn{3}{c}{\textbf{Near} ($15$\text{--}$35$\,units)} \\
            \cmidrule(lr){3-5} \cmidrule(lr){6-8}
             & & \textbf{F}$_{1\%}$ $\uparrow$ & \textbf{F}$_{5\%}$ $\uparrow$ & $\boldsymbol{\delta}_{1.05}$ $\uparrow$ & \textbf{F}$_{1\%}$ $\uparrow$ & \textbf{F}$_{5\%}$ $\uparrow$ & $\boldsymbol{\delta}_{1.05}$ $\uparrow$ \\
            \midrule
             & & -- & -- & -- & -- & -- & -- \\
            \checkmark & & $+5.43\%$ & $+4.19\%$ & $+4.22\%$ & $+3.51\%$ & $+2.81\%$ & $+2.86\%$ \\
            \checkmark & \checkmark & \hc{red!20}{$+10.15\%$} & \hc{red!20}{$+7.23\%$} & \hc{red!20}{$+7.23\%$} & \hc{red!20}{$+5.35\%$} & \hc{red!20}{$+4.69\%$} & \hc{red!20}{$+4.76\%$} \\
            \bottomrule
        \end{tabular}
    }
\end{table}

%% file: sec/8_conclusion.tex
% !TEX root = ../main.tex

\section{Conclusion}
\label{sec:conclusion}

We introduce \model{}, a training-free framework that enables streaming 3D reconstruction from arbitrarily long video under constant memory and compute.
By combining Self-Selective Caching with Anchor Protection, our method compresses the cache to a fixed budget while preserving geometrically critical tokens, achieving state-of-the-art accuracy across indoor, outdoor, and ultra-long sequence benchmarks with real-time throughput on a single consumer GPU.

\noindent\textbf{Limitations and Future Work.}
Despite operating under a fixed resource envelope, \model{} inherits the fundamental limitation of single-pass causal pipelines: geometric errors accumulate monotonically and cannot be corrected, as no mechanism exists for revisiting past predictions and each frame can only reference a bounded subset of prior context.
We believe staged streaming inference is a promising direction, combining mini-batch joint prediction with periodic lightweight global refinement to unite the bounded per-stage cost of causal models with the error-correction capacity of batch methods, mitigating long-horizon drift without full all-to-all recomputation.

%% file: sec/99_supp.tex
% !TEX root = ../main.tex

\newpage
\setcounter{section}{0}
\setcounter{figure}{0}
\setcounter{table}{0}
\renewcommand{\thesection}{\Alph{section}}
\renewcommand{\thefigure}{S\arabic{figure}}
\renewcommand{\thetable}{S\arabic{table}}

\begin{center}
  {\Large\bfseries Supplementary Material}
\end{center}

% =============================================================================
\section{Comparison with Full-Cache Baseline}
\label{sec:supp:fullcache}
% =============================================================================

\begin{figure*}[!hbp]
  \centering
  \includegraphics[width=\textwidth]{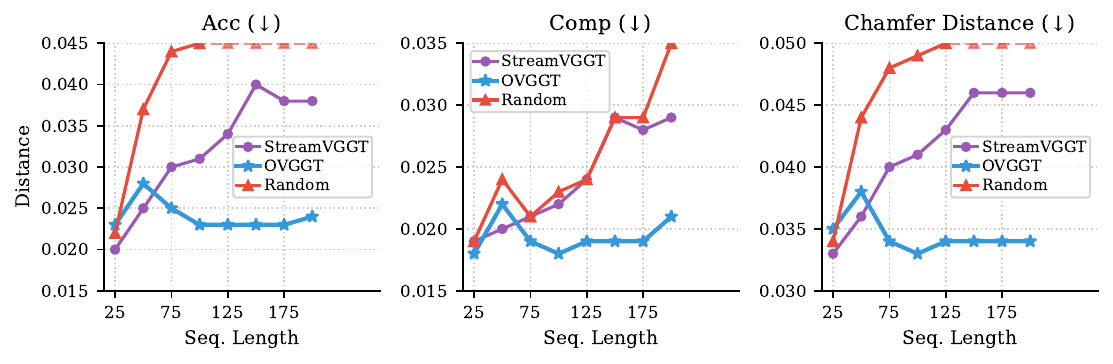}
  \caption{\textbf{Reconstruction quality vs.\ sequence length on 7-Scenes~\cite{shotton2013scenes}.}
  Mean Accuracy, Completeness, and Chamfer Distance are plotted from 25 to 200 frames.
  StreamVGGT (full cache) accumulates noise from redundant tokens as the sequence grows, causing progressive metric degradation.
  Random eviction under the same budget rapidly diverges, confirming that naive cache truncation is insufficient.
  \model{} selectively retains geometrically salient tokens, achieving lower and more stable error throughout.}
  \label{fig:drift_trend}
\end{figure*}

To provide a fine-grained view of how cache management affects reconstruction quality over time, we compare \model{} against the full-cache StreamVGGT~\cite{zheng2025streamvggt} and a random eviction baseline at sequence lengths from 25 to 200 frames on 7-Scenes~\cite{shotton2013scenes}.
StreamVGGT retains the entire KV cache up to its OOM limit (${\sim}200$ frames on 32\,GB), while the random baseline evicts tokens uniformly at random under the same $B{=}200\text{K}$ budget as \model{}.
\cref{fig:drift_trend} plots Accuracy, Completeness, and Chamfer Distance as a function of sequence length.

Two trends are evident.
First, StreamVGGT exhibits monotonically increasing error: as more tokens accumulate, the attention mechanism must attend over a growing pool of redundant and potentially noisy representations, diluting the effective contribution of geometrically informative tokens.
This observation directly supports the hypothesis that retaining the entire cache does not constitute an accuracy upper bound.
Second, random eviction rapidly diverges from both StreamVGGT and \model{}, demonstrating that the performance gain of \model{} is not merely a byproduct of cache size reduction but stems from the informed selection of which tokens to retain.
By contrast, \model{} maintains consistently low error across all sequence lengths, indicating that self-selective caching actively filters transient noise while preserving the critical geometric references needed for high-fidelity reconstruction.

% =============================================================================
\section{Camera Pose Estimation}
\label{sec:supp:pose}
% =============================================================================

Beyond dense 3D reconstruction, we also evaluate camera pose estimation quality by reporting Absolute Trajectory Error (ATE) on TUM-Dynamic~\cite{sturm2012benchmark} and ScanNet~\cite{dai2017scannet}.
Both datasets contain dynamic objects and complex camera motions that stress-test long-sequence tracking stability.

As shown in \cref{tab:camera_pose}, \model{} achieves the best ATE across nearly all sequence lengths on both datasets.
On TUM-Dynamic, our method consistently outperforms both Evict3R$^\dagger$ and InfiniteVGGT at every evaluation point from 100 to 1{,}000 frames, with the performance gap widening as sequence length increases.
This trend is particularly revealing: at 1{,}000 frames, \model{} reduces ATE by 30\% relative to Evict3R$^\dagger$ (0.058 vs.\ 0.083), confirming that dynamic anchor protection effectively suppresses the cumulative pose drift afflicting competing methods.

On ScanNet, where scenes are spatially larger and exhibit more diverse viewpoint changes, \model{} again leads at longer sequences.
InfiniteVGGT achieves a comparable ATE at 100 frames, and Evict3R$^\dagger$ achieves the best result at 300 frames; however, both degrade substantially beyond 500 frames.
These results confirm that self-selective caching combined with dynamic anchor protection yields more consistent camera trajectory estimation over extended sequences, complementing the 3D reconstruction gains reported in the main paper.

\input{table/quant_camera_pose}

% =============================================================================
\section{Camera Prediction Head}
\label{sec:supp:pose_head}
% =============================================================================

While the primary bottleneck lies in the aggregator's KV cache, StreamVGGT's camera prediction head also maintains its own KV cache that stores one token per frame.
Although the per-frame overhead is minimal (a single token versus $M{=}1{,}041$ in the aggregator), this cache still grows linearly with sequence length, incrementally degrading speed and violating the constant-cost property.

To ensure end-to-end $O(1)$ inference, we extend the same cache management framework (SSC + DAP) to the camera head.
Specifically, we allocate a separate budget to the camera head proportional to the number of frames that the total aggregator budget can accommodate.
Since each frame contributes only a single camera token, spatial smoothing is inapplicable and therefore omitted; all other components operate identically to the aggregator cache.
When DAP registers or demotes an anchor in the aggregator, the corresponding camera token is simultaneously protected or released in the camera head cache.
This synchronized management ensures that both the aggregator and the camera head operate under bounded memory, achieving truly constant-cost inference per frame across the entire pipeline.

% =============================================================================
\section{FFN Residuals as a Saliency Proxy}
\label{sec:supp:FFN_probing}
% =============================================================================

\begin{table*}[!htbp]
    \newcommand{\hc}[2]{{\setlength{\fboxsep}{1pt}\colorbox{#1}{#2}}}
    \centering
    \renewcommand{\arraystretch}{1.1}
    \setlength{\tabcolsep}{3pt}
    \caption{\textbf{Probing experiment: eviction strategy comparison.} 3D reconstruction on 7-Scenes~\cite{shotton2013scenes} at 100 and 300 frames under identical budget ($B{=}200\text{K}$) with DAP disabled. All strategies share the same hybrid scoring framework.}
    \label{tab:probing_eviction_updated}
    \resizebox{0.9\linewidth}{!}{
        \begin{tabular}{@{}l c cccc cccc@{}}
            \toprule
             & & \multicolumn{4}{c}{\textbf{100 frames}} & \multicolumn{4}{c}{\textbf{300 frames}} \\
            \cmidrule(lr){3-6} \cmidrule(l){7-10}
            \textbf{Strategy} & \textbf{FlashAttn} & Acc $\downarrow$ & Comp $\downarrow$ & NC $\uparrow$ & CD $\downarrow$ & Acc $\downarrow$ & Comp $\downarrow$ & NC $\uparrow$ & CD $\downarrow$ \\
            \midrule
            Attention weight            & \texttimes & 0.024 & 0.020 & 0.595 & 0.035 & \hc{red!20}{0.023} & 0.022 & \hc{red!20}{0.573} & \hc{red!20}{0.032} \\
            \midrule
            Q$\cdot$K dot product       & \checkmark & 0.026 & 0.020 & \hc{red!20}{0.596} & 0.035 & 0.032 & 0.022 & 0.570 & 0.037 \\
            \textbf{FFN residual (Ours)} & \checkmark & \hc{red!20}{0.023} & \hc{red!20}{0.018} & 0.595 & \hc{red!20}{0.033} & 0.026 & \hc{red!20}{0.020} & 0.571 & \hc{red!20}{0.032} \\
            \bottomrule
        \end{tabular}
    }
\end{table*}

A natural question is why the FFN residual magnitude serves as an effective geometric saliency proxy, and whether simpler or more direct alternatives would suffice.
To address this, we conduct controlled probing experiments on 7-Scenes at \textbf{both 100 and 300 frames} under the default budget ($B{=}200\text{K}$) with DAP disabled to isolate the effect of the scoring criterion.
We compare three eviction strategies that all share the same hybrid scoring framework, differing only in the current-frame scoring criterion:
(i)~attention-weight-based scoring, which materializes the full attention matrix to extract per-token importance;
(ii)~query-key dot product scoring ($\mathbf{q}\!\cdot\!\mathbf{k}$), which approximates attention-based importance without full materialization; and
(iii)~our FFN-residual-based activation scoring.
Crucially, only the latter two remain compatible with FlashAttention~\cite{dao2022flashattention,dao2024flashattention2}; attention-weight scoring requires materializing the $N{\times}N$ attention matrix, sacrificing memory efficiency and precluding fused attention kernels.

As shown in \cref{tab:probing_eviction_updated}, \textbf{our FFN-residual scoring demonstrates remarkable robustness across different sequence lengths.} 
At \textbf{100 frames}, it surprisingly outperforms even the attention-weight ``oracle'' across all metrics, suggesting that the FFN residual captures a more refined geometric signal than raw attention weights in the early stages of reconstruction. 
As the sequence extends to \textbf{300 frames}, while attention-weight scoring achieves slightly better Accuracy and Normal Consistency, it does so at the cost of FlashAttention incompatibility and significant memory overhead. 
Among the FlashAttention-compatible alternatives, our method consistently and substantially outperforms the $\mathbf{q}\!\cdot\!\mathbf{k}$ approximation. 
Notably, at 300 frames, FFN-residual scoring still matches the attention-weight oracle on Chamfer Distance and achieves the best overall Completeness.
The $\mathbf{q}\!\cdot\!\mathbf{k}$ dot product, despite being computationally lightweight, provides a noisier importance estimate that lacks the nonlinear refinement captured by the FFN.

These results confirm that FFN-residual scoring offers the most favorable trade-off: \textbf{it matches or even exceeds oracle-level reconstruction quality} while maintaining full FlashAttention compatibility and zero additional overhead. 
We attribute this effectiveness to the role of the FFN in geometric transformers.
The FFN applies a per-token nonlinear transformation that progressively refines raw visual features into geometrically grounded representations; tokens encoding structurally informative regions (\textit{e.g.}, edges, corners, depth discontinuities) undergo larger representational shifts, producing higher residual magnitudes.
This coarse-to-fine progression mirrors the feature hierarchy of vision transformers and provides a principled, zero-overhead importance signal naturally aligned with the demands of 3D reconstruction.

% =============================================================================
\section{Failure Cases}
\label{sec:supp:failure}
% =============================================================================

\begin{figure*}[!htbp]
  \centering
  \includegraphics[width=\textwidth]{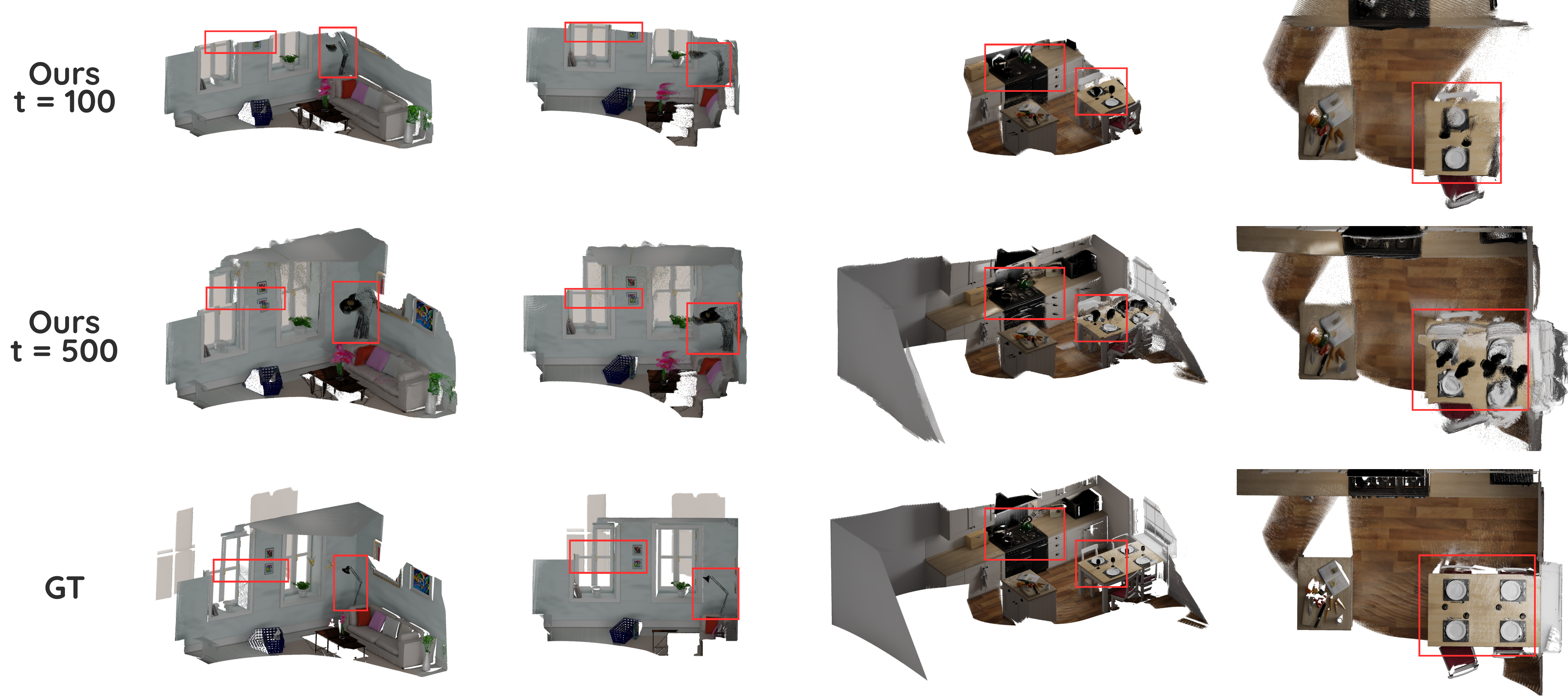}
  \caption{\textbf{Failure case analysis.} Reconstruction results of \model{} at $t{=}100$ and $t{=}500$ frames compared with the ground truth. Red boxes highlight regions where cumulative drift becomes apparent in longer sequences: later frames exhibit progressively degraded geometric fidelity, particularly at structural boundaries and fine-grained details.}
  \label{fig:failure}
\end{figure*}

While \model{} demonstrates robust performance across a wide range of scenarios, it inherits a fundamental limitation shared by single-pass pipelines: geometric errors accumulate monotonically along the sequence and cannot be rectified within a single inference pass. \cref{fig:failure} illustrates this effect on two representative indoor scenes. At $t{=}100$, the reconstructions closely match the ground truth in both global structure and local details. However, as the sequence extends to $t{=}500$, cumulative drift becomes increasingly visible. Structural boundaries exhibit misalignment, surface details degrade, and regions observed later in the sequence suffer disproportionately. This is because these regions rely entirely on a cache that has undergone repeated compression cycles without a mechanism for revisiting or correcting earlier predictions.

This limitation is intrinsic to the strictly forward streaming paradigm. Since each frame is processed exactly once and past predictions are never revisited, small per-frame errors compound over time without the possibility of global correction. Although our design preserves geometrically critical tokens and suppresses coordinate-system inconsistency, it still operates under the constraint that information flows exclusively forward in time. This deviates from human spatial intuition, where prior estimates and future expectations are dynamically updated based on current sensory inputs to maintain a globally consistent internal map.

As discussed in conclusion, we posit that adopting a staged streaming inference approach represents a promising direction to address this limitation. This could evolve in two primary directions: (1) \textbf{Mini-batch global estimation}, which utilizes a sliding window of multiple frames to perform joint prediction, thereby mitigating information scarcity; and (2) \textbf{Periodic lightweight global refinement}, which employs triggered global optimization to rectify past estimates and ensure long-term consistency. Such strategies would significantly alleviate drift in long-range sequences while remaining manageable and intuitive in terms of resource allocation and cache management.

% =============================================================================
\section{Baseline Adaptations}
\label{sec:supp:baseline_modi}
% =============================================================================

To ensure fair and complete evaluation across all sequence lengths, we applied the following adaptations to baseline methods whose official implementations cannot natively handle long sequences or accurately reflect model-only VRAM usage.
All modifications preserve the original model weights and inference logic; only the data loading and batching strategy is changed.

\noindent\textbf{Spann3R~\cite{wang2024spann3r}.}
We use the original forward function, which processes consecutive frame pairs $(t{-}1, t)$ sequentially.
All input frames are retained in CPU memory and transferred to GPU only when actively needed.

\noindent\textbf{CUT3R~\cite{wang2025cut3r}.}
The official pipeline loads all input frames onto the GPU simultaneously for batch encoding before performing sequential decoding.
This causes VRAM to scale linearly with input length until OOM, preventing evaluation on long sequences.
We restructured the inference pipeline to encode and decode each frame sequentially, ensuring stable operation across all tested sequence lengths without altering the model itself.

\noindent\textbf{Point3R~\cite{wang2025point3r}.}
Point3R shares the same batch-encoding bottleneck.
To balance throughput and memory, we adopt a chunked inference strategy: input frames are encoded in chunks of 10, followed by per-chunk decoding.
This allows Point3R to process sequences of up to 1{,}000 frames without OOM while maintaining reasonable efficiency.

For all baselines, input data resides in CPU memory and is moved to GPU on demand, so that all reported VRAM figures accurately reflect model inference costs rather than data staging overhead.

% =============================================================================
\section{Ultra-Long Sequence Evaluation Details}
\label{sec:supp:ultra_long_setting}
% =============================================================================

The Long3D~\cite{yuan2026infinitevggt} dataset contains sequences of up to 10{,}000 consecutive frames, producing dense point clouds of unprecedented scale.
At this scale, standard point cloud registration pipelines become computationally prohibitive and prone to failure due to the sheer volume of points and the large spatial extent of the reconstructions.
We therefore implement a robust multi-stage evaluation pipeline.
First, the raw dense point clouds are downsampled to a tractable resolution.
Second, Statistical Outlier Removal (SOR) is applied to suppress noise and spurious points.
Third, adaptive DBSCAN clustering segments the point cloud into coherent spatial clusters, with the clustering scale adapted to the spatial extent of each reconstruction.
Fourth, feature-based RANSAC coarse registration aligns the predicted and ground-truth point clouds.
Finally, a two-stage ICP (coarse-to-fine) refinement produces the final alignment from which all metrics are computed.

% =============================================================================
\section{Effect of Maximum Anchor Count}
\label{sec:supp:anchor_kmax}
% =============================================================================

\begin{table}[]
    \newcommand{\hc}[2]{{\setlength{\fboxsep}{1pt}\colorbox{#1}{#2}}}
    \centering
    \renewcommand{\arraystretch}{0.9}
    \setlength{\tabcolsep}{4pt}
    \caption{{Effect of maximum historical anchor count $K_{\max}$.} Video depth estimation on KITTI~\cite{geiger2013vision} at 500 frames under $B{=}200\text{K}$. \hc{red!20}{Best} results highlighted.}
    \label{tab:ablation_kmax}
    \resizebox{0.8\linewidth}{!}{
        \begin{tabular}{@{}c cc cc@{}}
            \toprule
            \multirow{2}{*}{$K_{\max}$} & \multicolumn{2}{c}{{Near}} & \multicolumn{2}{c}{{Far}} \\
            \cmidrule(lr){2-3} \cmidrule(lr){4-5}
            & RMSE $\downarrow$ & $\delta_{1.25}$ $\uparrow$ & $\delta_{1.05}$ $\uparrow$ & F$_{1\%}$ $\uparrow$ \\
            \midrule
            1  & $1.553$ & $0.861$ & $0.091$ & $0.018$ \\
            \textbf{3} & \hc{red!20}{$1.546$} & \hc{red!20}{$0.863$} & \hc{red!20}{$0.092$} & \hc{red!20}{$0.019$} \\
            5  & $1.552$ & $0.862$ & $0.091$ & $0.018$ \\
            10 & $1.548$ & \hc{red!20}{$0.863$} & $0.091$ & $0.018$ \\
            \bottomrule
        \end{tabular}
    }
\end{table}

\cref{tab:ablation_kmax} reports depth estimation metrics on KITTI~\cite{geiger2013vision} over 500 frames with a default budget of $B=200\text{K}$, stratified into Near ($d < 35$ units) and Far ($d > 35$ units) depth ranges.
While enabling versus disabling anchor protection yields a substantial accuracy gap, the sensitivity to the exact number of active anchors is comparatively mild once protection is enabled.

With $K_{\max}{=}1$, only a single historical anchor is available, which may become spatially distant from the current view during extended traversals, leading to reduced depth accuracy.
Increasing to $K_{\max}{=}3$ yields consistent improvement across all metrics, as the three active anchors collectively cover a wider portion of the trajectory, providing sufficient long-range geometric references to suppress drift in both near and far ranges.
However, further increasing $K_{\max}$ to 5 or 10 brings negligible additional gain and can even cause marginal degradation, since each additional anchor consumes $\lceil\eta \cdot N_p\rceil$ protected tokens per layer, reducing the evictable pool capacity and thereby limiting the distributional diversity that the hybrid scoring mechanism relies upon.
Beyond a modest threshold, the marginal benefit of additional anchors is outweighed by the loss of flexible cache capacity.
We therefore adopt $K_{\max}{=}3$ as the default, providing ample geometric anchoring while preserving sufficient budget for the evictable pool to maintain broad scene coverage.

%% file: table/quant_camera_pose.tex
% ==========================================
% Table: Camera Pose (ATE only, seq len horizontal)
% ==========================================
\begin{table*}[t]
    \newcommand{\hc}[2]{{\setlength{\fboxsep}{1pt}\colorbox{#1}{#2}}}
    \centering
    \caption{Camera Pose evaluation (ATE $\downarrow$) on TUM-Dynamic~\cite{sturm2012benchmark} and ScanNet~\cite{dai2017scannet} across different sequence lengths. \hc{red!20}{Best} results highlighted.}
    \label{tab:camera_pose}
    \renewcommand{\arraystretch}{1.05}
    \setlength{\tabcolsep}{4pt}
    \resizebox{0.9\linewidth}{!}{
        \begin{tabular}{l ccccc ccccc}
            \toprule
            \multirow{2}{*}[-3pt]{Method} & \multicolumn{5}{c}{TUM-Dynamic} & \multicolumn{5}{c}{ScanNet} \\
            \cmidrule(lr){2-6} \cmidrule(lr){7-11}
            & 100 & 300 & 500 & 800 & 1k & 100 & 300 & 500 & 800 & 1k \\
            \midrule
            Evict3R$^\dagger$~\cite{mahdi2025evict3r}
                & $0.017$ & $0.032$ & $0.045$ & $0.074$ & $0.083$
                & $0.060$ & \hc{red!20}{$0.184$} & $0.285$ & $0.430$ & $0.491$ \\
            InfiniteVGGT~\cite{yuan2026infinitevggt}
                & $0.018$ & $0.034$ & $0.046$ & $0.061$ & $0.066$
                & \hc{red!20}{$0.056$} & $0.202$ & $0.329$ & $0.438$ & $0.478$ \\
            \textbf{Ours}
                & \hc{red!20}{$0.014$} & \hc{red!20}{$0.029$} & \hc{red!20}{$0.039$} & \hc{red!20}{$0.047$} & \hc{red!20}{$0.058$}
                & \hc{red!20}{$0.056$} & $0.198$ & \hc{red!20}{$0.284$} & \hc{red!20}{$0.385$} & \hc{red!20}{$0.427$} \\
            \bottomrule
        \end{tabular}
    }
\end{table*}